\documentclass[twocolumn,twoside,letter]{IEEEtran}
\usepackage{times}
\usepackage{amsmath}
\usepackage{amsthm}
\usepackage{amssymb}
\usepackage{graphicx}
\usepackage{subcaption}
\usepackage{verbatim}
\usepackage{tabularx}
\usepackage[thinlines,thiklines]{easybmat}
\usepackage{latexsym}
\usepackage[numbers]{natbib}
\usepackage{multicol}
\usepackage[bookmarks=true]{hyperref}
\usepackage{stmaryrd}
\usepackage{multirow}
\usepackage{textcomp}
\usepackage{diagbox}
\usepackage{array}
\usepackage{color,soul}
\usepackage{float}
\usepackage[linesnumbered,ruled,vlined]{algorithm2e}
\usepackage[noend]{algpseudocode}
\algrenewcommand\algorithmiccomment[1]{\hfill $\triangleright$ \textit{\footnotesize #1}}
\usepackage{tikz}
\usetikzlibrary{arrows.meta, positioning, calc, decorations.markings, 3d}
\usepackage{pgfplots}

\usepackage{array}
\usepackage{booktabs}

\sethlcolor{yellow} 

\newcolumntype{M}[1]{>{\centering\arraybackslash}m{#1}}
\newcolumntype{N}{@{}m{0pt}@{}}

\usepackage[left=0.75in,top=0.75in,right=0.75in,bottom=0.75in]{geometry}
\newbox\tempbox

\newcommand{\bs}[1]{\ensuremath{{\boldsymbol{#1}}}}

\newenvironment{pf}{\noindent {\bf Proof:} }{\qed \\}
\newtheorem{thm}{Theorem}

\newtheoremstyle{remark}
  {} 
  {} 
  {} 
  {} 
  {\bfseries} 
  {.} 
  { } 
  {} 

\theoremstyle{remark}

\newtheoremstyle{corollary}
  {} 
  {} 
  {} 
  {} 
  {\bfseries} 
  {.} 
  { } 
  {} 

\theoremstyle{corollary}
\newtheorem{corollary}{Corollary}

\newtheoremstyle{observation}
  {} 
  {} 
  {} 
  {} 
  {\bfseries} 
  {.} 
  { } 
  {} 

\theoremstyle{observation}

\newtheoremstyle{lemma}
  {} 
  {} 
  {} 
  {} 
  {\bfseries} 
  {.} 
  { } 
  {} 

\theoremstyle{lemma}
\newtheorem{lemma}{Lemma}
\newtheoremstyle{lemma}
  {} 
  {} 
  {} 
  {} 
  {\bfseries} 
  {.} 
  { } 
  {} 

\theoremstyle{proof}

\begin{document}
\title{Optimal Planning for Multi-Robot \\ Simultaneous Area and Line Coverage Using Hierarchical Cyclic Merging Regulation}
\author{Tianyuan Zheng, Jingang Yi, and Kaiyan Yu~\thanks{T. Zheng and J. Yi are with the Department of Mechanical and Aerospace Engineering, Rutgers University, Piscataway, NJ 08854 USA (email: tz270@scarletmail.rutgers.edu; jgyi@rutgers.edu). K. Yu is with the Department of Mechanical Engineering, Binghamton University, Binghamton, NY 13902 USA (email: {kyu@binghamton.edu}).}}

\maketitle

\begin{abstract}
The double coverage problem focuses on determining efficient, collision-free routes for multiple robots to simultaneously cover linear features (e.g., surface cracks or road routes) and survey areas (e.g., parking lots or local regions) in known environments. In these problems, each robot carries two functional roles: service (linear feature footprint coverage) and exploration (complete area coverage). Service has a smaller operational footprint but incurs higher costs (e.g., time) compared to exploration. We present optimal planning algorithms for the double coverage problems using hierarchical cyclic merging regulation (HCMR). To reduce the complexity for optimal planning solutions, we analyze the manifold attachment process during graph traversal from a Morse theory perspective. We show that solutions satisfying minimum path length and collision-free constraints must belong to a Morse-bounded collection. To identify this collection, we introduce the HCMR algorithm. In HCMR, cyclic merging search regulates traversal behavior, while edge sequence back propagation converts these regulations into graph edge traversal sequences. Incorporating balanced partitioning, the optimal sequence is selected to generate routes for each robot. We prove the optimality of the HCMR algorithm under a fixed sweep direction. The multi-robot simulation results demonstrate that the HCMR algorithm significantly improves planned path length by at least 10.0\%, reduces task time by at least 16.9\% in average, and ensures conflict-free operation compared to other state-of-the-art planning methods.
\end{abstract}

\IEEEpeerreviewmaketitle

\section{Introduction}
\label{Sec:intro}

\begin{figure}[t!]
	\centering
	\includegraphics[trim=20cm 8cm 8cm 5cm, clip,width=1.04\linewidth]{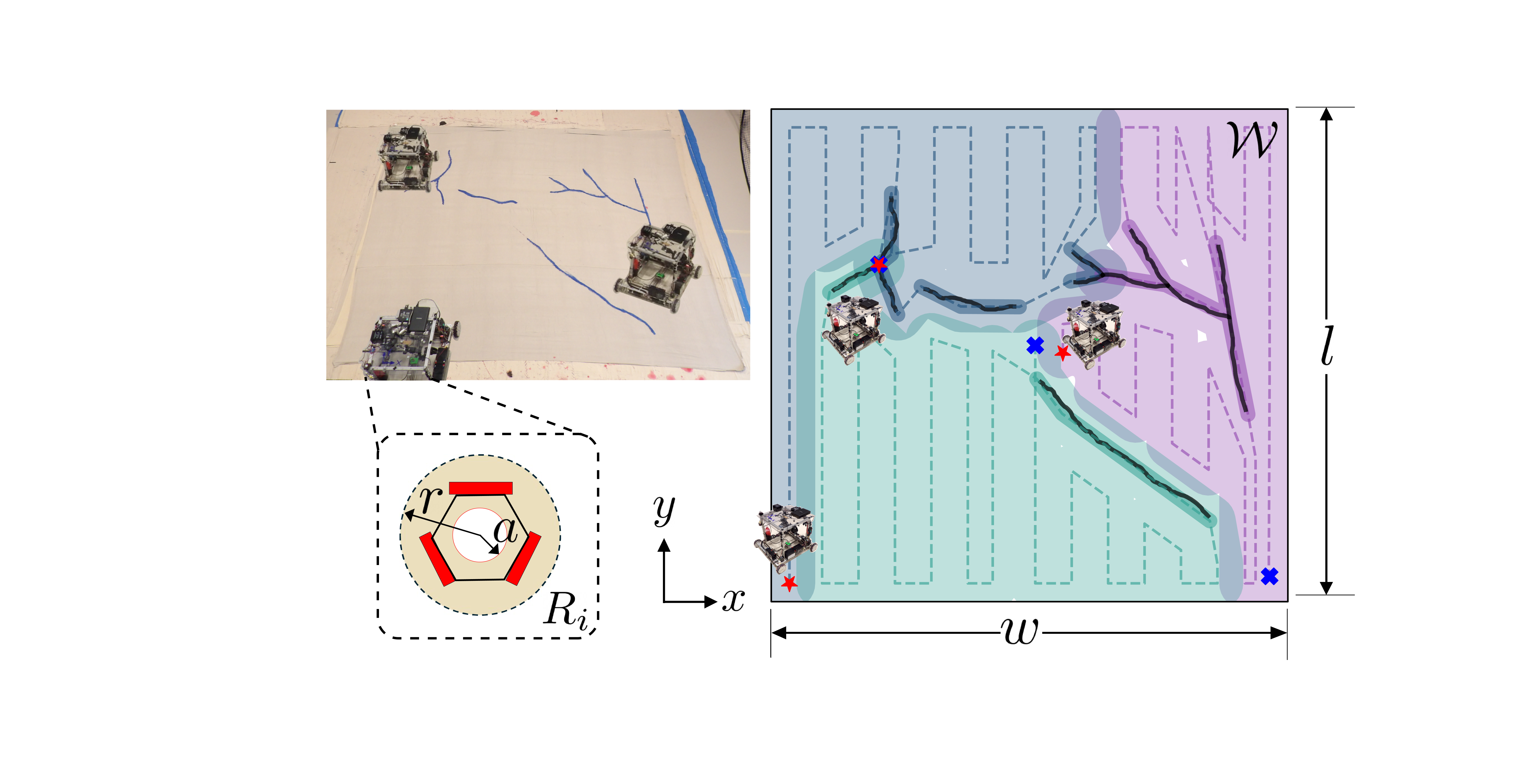}
	\caption{An example of the MDC problem. The top-left image illustrates a real-world experiment, where blue curves represent linear features (cracks) to be covered, and three robots coordinate to perform the task. The bottom-left image shows each robot is based on an omni-wheel system, and equipped with a wide sensing region and a filling footprint. The right-side image presents the corresponding simulation on the real-world map, where black bold curves represent linear features. A robotic team \( \mathcal{R} \) is deployed to fill the cracks while ensuring complete coverage of \( \mathcal{W} \). In this example, \( N=3 \), with robot starting and ending points marked as red stars and blue crosses, respectively. Different colors indicate the trajectories of individual robots.}

	\label{fig:crack_filling_illu}
\end{figure}

Area coverage and line coverage have emerged as key challenges in a wide range of robotic applications. Area coverage focuses on surveying open spaces to ensure full visibility and detect various objects of interest~\cite{almadhoun2019survey}, supporting tasks in environmental monitoring~\cite{hitz2017adaptive}, infrastructure inspection~\cite{matlekovic2023constraint}, agricultural surveying~\cite{hoffmann2023optimal}, and search-and-rescue operations~\cite{cho2021coverage}. In contrast, line coverage centers on the careful inspection or repair of linear features, such as surface cracks, pipelines, railway lines, and utility lines~\cite{agarwal2024line,su2023balanced}. A practical example of combined area and line coverage requirements is a bridge deck inspection~\cite{GucunskiIJIRA2017}, as illustrated in Fig.~\ref{fig:crack_filling_illu}. In this scenario, an aerial drone first surveys the region to gather preliminary information about the distribution of surface cracks across a large area. Based on these data, multiple ground robots equipped with cameras, non-destructive evaluation sensors~\cite{LaTMech2013}, and material filling nozzles are deployed. These robots perform two tasks simultaneously: constructing detailed feature maps of the deck, such as delamination and subsurface crack maps (area coverage), and following identified cracks to repair them in real time (line coverage). 

Inspired by the above-mentioned applications, we formulate the double coverage problem as a unified framework that integrates area coverage and linear feature servicing into a single optimization task. For multi-robot systems, this is generalized to the multi-double coverage (MDC) problem, where each robot conducts two functional roles: exploration for scanning regions and service for filling or repairing linear features such as surface cracks. In addition, robots adhere to collision-free constraints, ensuring that their exploration paths do not intersect with those of their teammates. This study assumes that linear feature locations are known a priori, and the results presented here serve as a stepping stone for future development of online algorithms for unknown environments.

The MDC planning is closely related to the $k$-covering salesman problem, a variant of the $k$-traveling salesman problem (TSP), where multiple agents collectively visit all specified cities while minimizing the total distance. In this context, we specifically denote two variants of the TSP: the Chinese Postman Problem (CPP) and the rural postman problem. While TSP and $k$-TSP focus solely on visiting discrete sets of points (cities) on a given graph, the MDC problem involves transitioning from the continuous physical environment, where line features form intricate patterns, to an appropriate graph-based abstraction. This transition introduces subtleties not present in standard TSP formulations. In particular, even the shortest routes within the extracted graph may not directly translate into the true shortest trajectories for robots operating in the actual workspace. To achieve truly optimal paths, the MDC approach must bridge the gap between the topological complexities of the linear feature map and its graph representation. Additionally, to ensure efficient and fair utilization of robotic resources, the MDC problem requires balanced partitioning of tasks among multiple robots.

It is worth noting that all these problems are NP-hard. While efficient approximation algorithms with theoretical bounds exist for the \( k \)-TSP problem, the gap introduced above invalidates these bounds for the MDC problem. To address this, we use Morse topological analysis to develop the  hierarchical cyclic merging regulation (HCMR) heuristic algorithm. The HCMR reduces the search space to paths where each edge is Morse bounded. This approach makes finding an optimal solution for the MDC problem feasible. We evaluate our method through detailed numerical simulations across varying numbers of robots and map densities, comparing it against state-of-the-art baselines to validate our design choices. Results demonstrate that HCMR consistently and significantly outperforms all baselines, improving planned path length by at least $10.0$\%, reducing task time by at least $16.9$\% in average, and also ensuring conflict-free operation. The main contributions are summarized as follows.
\begin{enumerate}
    \item We introduce a novel MDC problem formulation that unifies area exploration and linear servicing into a single framework. This includes well-defined objective functions and constraints that address practical application requirements.  
    \item We establish a connection between topology and graph theory by quantifying discontinuous behavior in edge connections within the manifold space and formulating regulation rules in the graph space.
    \item We propose the HCMR algorithm to effectively reduce the search space by identifying a Morse-bounded collection, enabling the discovery of complete coverage and optimal paths within a manageable subset of all possible solutions.  
    \item Building on the Morse-bounded set identified by the HCMR algorithm, we develop a method to generate locally optimal, collision-free trajectories for each robot.
\end{enumerate}

\section{Related Work}
\label{Sec:relate}

In this section, we position our contributions within relevant subfields, emphasizing the similarities to and distinctions from prior research efforts.

\subsection{Area Coverage}

Area coverage has long been recognized as an NP-hard problem~\cite{arkin2000approximation} and has been extensively studied~\cite{galceran2013survey,cabreira2019survey}. Solutions to this problem are typically categorized into grid-based and exact methods. Grid-based methods such as~\cite{gabriely2001spanning,wei2018coverage} discretized the environment at a predefined resolution, simplifying problem transformation and enabling learning-based approaches~\cite{apuroop2021reinforcement}. However, their effectiveness depends heavily on the resolution, making them less suitable for environments with complex geometries.

In this paper, we concentrate on exact methods for complete area coverage, specifically Morse-based cellular decomposition and generalized Boustrophedon decomposition (BCD)~\cite{howieIJRR2002}. These techniques consider topological changes at critical points to divide the environment into manageable cells. A critical point is located on the boundary of an object whose surface normal is perpendicular to the sweep direction. In~\cite{rekleitis2008efficient,xu2014efficient}, a Reeb graph was constructed from the BCD, where cells were represented as edges and their connectivity as vertices. An algorithm designed for CPP computes an Eulerian tour on this Reeb graph, delineating a sequence of cells for robotic navigation. However, the optimality of the planned paths using these methods cannot be guaranteed. The inherent back-and-forth planning strategies within cells and transitions between them often result in suboptimal paths. Furthermore, not all Eulerian tours are equivalent in terms of efficiency and coverage effectiveness. While~\cite{mannadiar2010optimal} proposed an optimal strategy based on a specific Eulerian tour, it failed to account for variations in robot paths resulting from different Eulerian tours.

Building on the Reeb graph framework, the work in~\cite{karapetyan2017efficient} extended to multi-robot systems using clustering and the $k$-CPP algorithm. Similarly, the approach in~\cite{agarwal2022area} optimized the number of turns for multiple robots by relaxing the BCD to non-monotone polygons and allowing non-unified sweep directions within cells. However, these methods only consider analyses to a single Eulerian tour for task partitioning, potentially missing further efficient distributions of tasks among robots.
\vspace{-1mm}
\subsection{Line Coverage}
\vspace{-1mm}
Line coverage problems focus on deploying agents to cover linear features using their sensor footprints. Graph representations of these features are typically simplified or modified based on sensor footprint sizes. In~\cite{oh2014coordinated}, road network coverage was formulated as a mixed integer linear programming problem. A nearest insertion heuristic was combined with the CPP algorithm to solve the single-robot case, and an auction algorithm is used to distribute tasks among multiple robots. While $k$-means clustering was commonly applied for task allocation, the work in~\cite{agarwal2024line} introduced a merge-embed-merge heuristic to improve allocation by considering turning costs and nonholonomic constraints in large graphs. Notably, our problem generalizes line coverage by treating non-required edges connecting linear features as Reeb-graph edges, which are then treated as required for coverage.

Research on double coverage tasks involving robots is limited. In~\cite{kaiyanTRO2024,yu2019icra}, near-optimal solutions for a single robot performing double coverage were explored in both known and unknown environments. Building on this foundation, we address the problem of finding optimal solutions for multiple robots in known environments. Additionally, our work draws connections to 4-regular graph theory and $\kappa$-transformation~\cite{traldi2012interlacement}. The estimation of Eulerian tours using 4-regular graph theory~\cite{tutte1941unicursal} motivates our approach to reducing the vast search space. Meanwhile, $\kappa$-transformation provides the theoretical basis for the HCMR heuristics introduced in this paper.

\section{Problem Formulation} 

\label{Sec:pbstate}

\subsection{Notations and Preliminaries}

We use $\mathbb{G}=(\bs{V},\bs{E})$ to represent a general graph, where $\bs{V}$ is the vertex set and $\bs{E} \subset \bs{V} \times \bs{V}$ is the edge set. For $\mathbb{G}$, let $\bs{S} \subset \bs{V}$ be a subset of $\bs{V}$, $\mathbb{G}/\bs{S}$ denotes {\em vertex contraction} operation and $\mathbb{G}[\bs{S}]$ denotes {\em induced graph} of $\mathbb{G}$. A connected component is a maximal subgraph in which any two vertices are connected to each other by a path within the subgraph. We call an edge $B \in \bs{E}$ a \textit{bridge} if $\mathbb{G}-B$ has more connected components than $\mathbb{G}$. A path $p$ consisting of at least three edges is called a \textit{cycle} if it is closed (i.e., the starting and ending vertices are the same) and no vertices (except for the starting/ending vertex) or edges are repeated. The cycle and bridge spaces, denoted by sets $\bs{E}_\text{bridge}$ and $\bs{E}_\text{cycle}$, are the set of all cycles and bridges in $\mathbb{G}$, respectively. We have $\bs{E}=\bs{E}_\text{bridge} \cup \bs{E}_\text{cycle}$ and $\bs{E}_\text{bridge} \cap \bs{E}_\text{cycle} = \emptyset$. Two cycles are called vertex/edge \textit{adjacent} if they share at least one vertex/edge. Let ${C}_1$ and ${C}_2$ be two cycles in $\mathbb{G}$, \textit{symmetric difference} for them is defined as ${C}_1 \triangle {C}_2=({C}_1 \setminus {C}_2) \cup ({C}_2 \setminus {C}_1)$, that is, combining two cycles into one larger cycle by removing the edges that are common to both cycles and retaining only the edges that are unique to each cycle. A cycle basis $\mathcal{K}$ is a set of cycles such that any cycle in the cycle space can be expressed as a symmetric difference of cycles in $\mathcal{K}$. 

In the topological sense, for a Euclidean space or, more generally, a manifold $X$, the interior and boundary of $X$ are denoted by $\mathrm{int}(X)$ and $\partial(X)$, respectively. $\mu (\cdot)$ represents the Lebesgue measure of a subset of $X$. A function \( f: X \to Y \) is \textit{almost everywhere continuous} (a.e. continuous) if the set of discontinuities has Lebesgue measure zero, i.e., \( \mu(\{ x \in X \mid f \text{ is not continuous at } x \}) = 0 \). A \textit{quotient space} \( X / \sim \) (also denoted as \( X / q \), where \( q: X \to X / q \) is the quotient map) is formed by contracting equivalent points in \( X \) according to an equivalence relation \( \sim \). The resulting space consists of equivalence classes as single points. Given two manifolds \( \mathcal{M}_1 \) and \( \mathcal{M}_2 \), their \textit{disjoint union}, denoted as \( \mathcal{M}_1 \sqcup \mathcal{M}_2 \), is the space consisting of all points in \( \mathcal{M}_1 \) and \( \mathcal{M}_2 \) with the induced topology, treating them as separate, non-interacting components.

\subsection{MDC Problem Formulation}

Without loss of generality, the robot exploration is considered in a rectangular area \(\mathcal{W} \subset \mathbb{R}^2\) of size \(l \times w\). Fig.~\ref{fig:crack_filling_illu} illustrates an example of robotic crack inspection and filling task. In this case, the linear feature is considered as a set of surface cracks \(\mathcal{C} \subset \mathcal{W}\). A team of $N \in \mathbb{N}$ robots, denoted by \(\mathcal{R} = \{ R_1, R_2, \ldots, R_{N} \}\), is deployed to survey area $\mathcal{W}$ while filling these cracks $\mathcal{C}$. Initially, the robots navigate to predetermined optimal positions without performing any tasks, thereby disregarding the associated positioning costs. While the objective for $\mathcal{R}$ is to collaboratively cover \(\mathcal{W}\) to detect and simultaneously fill all cracks. In this paper, we focus on the foundational scenario where the locations of $\mathcal{C}$ are given and known.

Each robot operates in two task modes: exploration $T_e$ (scanning the area) and service $T_s$ (tracking linear features, e.g., filling cracks). Each robot is equipped with a circular service footprint area $\mathcal{F}$ with radius \(a\), and an exploring perception area $\mathcal{H}$ with radius \(r\), where \(r \gg a\). Let \(\pi_i\) denote the trajectory for robot \(R_i\) and for any point $x \in \pi_i$, robot $R_i$ conducts task $T_x$. The MDC constraint is represented by
\begin{equation}
 \mathop{\bigcup}\limits_{i=1}^{N} \mathop{\bigcup}\limits_{x \in \pi_i} (x \oplus \mathcal{H}) \supset \mathcal{W}, \;\,  \mathop{\bigcup}\limits_{i=1}^{N} \mathop{\bigcup}\limits_{x \in \pi_i} (x \oplus \mathcal{F}) \mathbb{I}(T_x=T_s) \supset \mathcal{C},
\label{eq.linear_coverage}
\end{equation}
where \(\oplus\) represents the dilation calculation, i.e., the Minkowski sum. $\mathop{\bigcup}_{x \in \pi_i} (x \oplus \mathcal{H})$ gives the area by dilating path $\pi_i$ by the perception area $\mathcal{H}$. The function \( \mathbb{I}(T_x = T_j) \) represents the \textit{indicator function}, which equals 1 if \( T_x = T_j \) and 0 otherwise, where \( j \in \{e, s\} \). The indicator function can also represent other events in a binary manner.  
 To simplify the switching complexity between tasks $T_e$ (exploration) and $T_s$ (service), similar to~\cite{kaiyanTRO2024,yu2019icra,guo2017optimal}, we assume that the robot's service motion $v_s$ is generally much faster than its exploration velocity $v_e$. Consequently, the time required for completing $T_s$ is significantly shorter than that for $T_e$. Based on this assumption, we primarily focus on the exploration task while incorporating collision-free path constraints, as the impact of service motion on overall planning is minimal. Moreover, finding routes that are conflict-free across robots is challenging \cite{ren2023cbss,sharon2015conflict}. Therefore, we consider following motion constraints. 
\begin{gather}
 \Bigl(\mathop{\bigcup}\limits_{x \in \pi_i} x \mathbb{I}(T_x=T_e) \Bigr) \bigcap \Bigl( \mathop{\bigcup}\limits_{x' \in \pi_j} x' \mathbb{I}(T_{x'}=T_e)\Bigr)= \emptyset, \label{eq.area_collision_constraint} \\
 \mu \Bigl( \bigl(\mathop{\bigcup}\limits_{x \in \pi_i} x \mathbb{I}(T_x=T_s)\bigr) \bigcap \bigl(\mathop{\bigcup}\limits_{x' \in \pi_j} x' \mathbb{I}(T_{x'}=T_s) \bigr) \Bigr)= 0, \label{eq.crack_collision_constraint}
\end{gather}
for $ 1 \leq i \neq j \leq N$. Constraint~(\ref{eq.area_collision_constraint}) ensures that the exploration paths of different robots are strictly disjoint, while~(\ref{eq.crack_collision_constraint}) allows intersections only at points where crack branches intersect, thereby promoting disjoint paths otherwise.

A fully connected network among $\mathcal{R}$ is assumed so that all robots maintain full information $\mathcal{C}$ and decisions are made in a centralized manner. We also assume that the robots follow holonomic kinematics as part of its configuration. This assumption simplifies motion planning by allowing the robots to move in any direction with free rotation. Furthermore, location of $R_i$ is precisely known at all times. The planning objective is to determine $\bs{\pi}_R=\{\pi_1,\cdots,\pi_N\}$ to minimize the total length and the maximum time required for robots to complete the MDC, that is, 
\begin{equation}
\label{eq.objective}
\min\limits_{\bs{\pi}_R} \; \Bigl( \sum_{i=1}^{N} L(\pi_i) + \alpha \max\limits_{1\leq i \leq N} t_{f_i}\Bigr),
\end{equation}
where $L(\pi_i)$ is the distance function of $\pi_i$, $\alpha >0$ is the weight factor, and $t_{f_i}$ is the finishing time for $R_i$, $1\leq i \leq N$.



\begin{figure*}[h!]
     \centering

    \includegraphics[trim=0cm 7cm 0.5cm 0cm, clip,width=\textwidth]{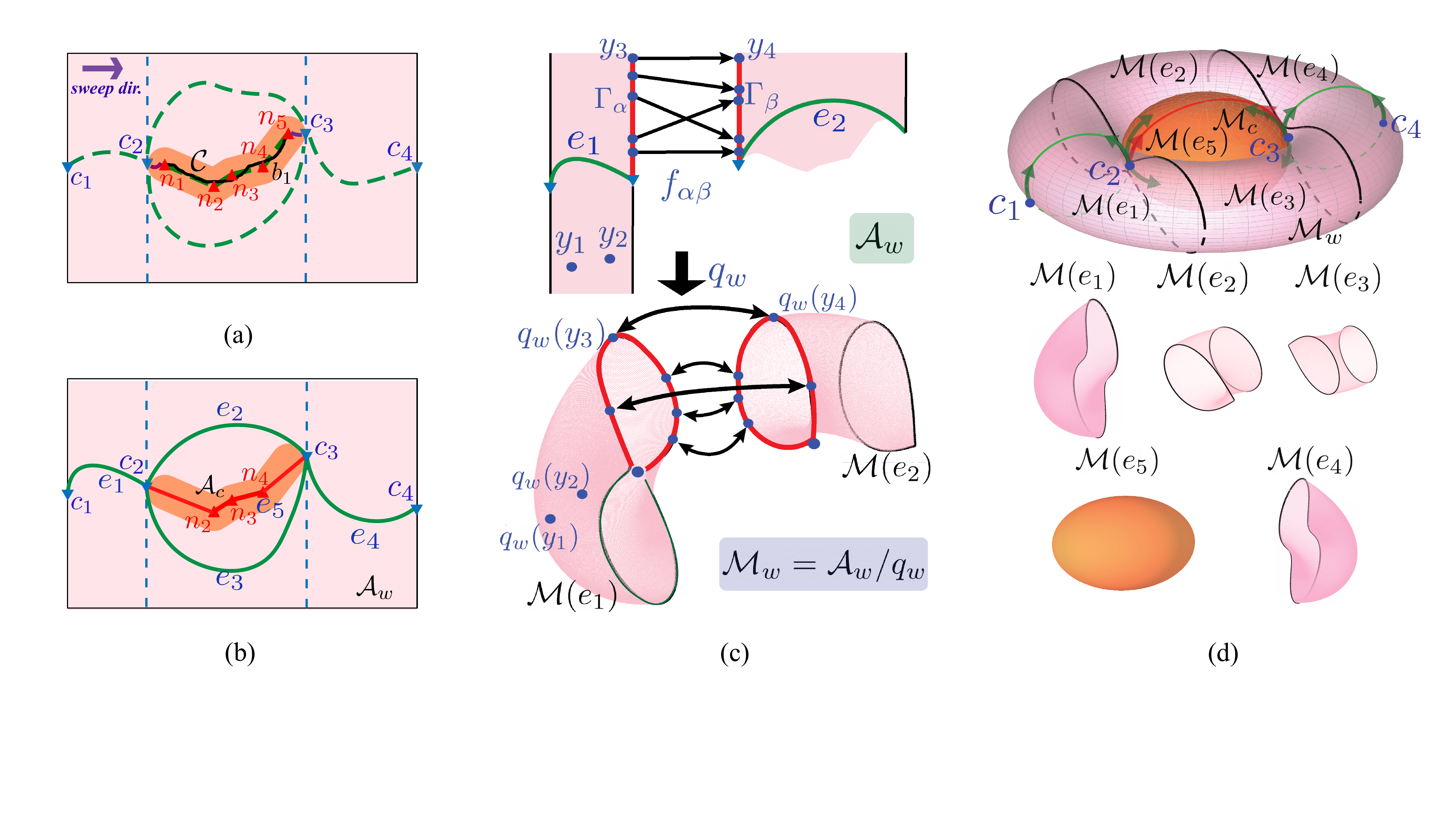}

\caption{
An example of graph construction and its relationship with Morse theory, illustrated using a toy model with a single branch. 
(a) The augmented graph, \(\mathbb{G}_\text{aug}\), is constructed by integrating the Reeb graph and the crack graph. 
(b) The simplified graph, \(\mathbb{G}_\text{sim}\), is obtained by contracting \(\mathbb{G}_\text{aug}\) at critical points. 
(c) Construction of the Reeb manifold cells corresponding to \(e_1\) and \(e_2\) in \(\mathbb{G}_{\text{sim}}\) using the quotient map \(q_w\). 
(d) Top shows two manifolds, \(\mathcal{M}_w\) and \(\mathcal{M}_c\), corresponding to \(\mathcal{A}_w\) and \(\mathcal{A}_c\), respectively. Below shows the decomposed manifold cells.
}

     \label{fig.graph_construction}
\end{figure*}

            



\section{Morse Boundedness} \label{Sec:morsetheory}
In this section, we first present the method for constructing graphs. We then discuss the necessity of regulation in generating Eulerian tours on these graphs. From the perspective of Morse theory in manifold space, we propose that the set of Eulerian tour candidates should form a Morse-bounded collection to ensure well-structured and efficient tour generation.


\subsection{Linear Graph and Morse Decomposition}
We present the graph construction method used in this paper and illustrate it with a toy model, as shown in Fig.~\ref{fig.graph_construction}.
We first partition $\mathcal{C}$ into distinct branches. A branch $\bs{b}$ is defined as a set of points, in which every point satisfies 
\begin{equation}\label{eq.branchdef}
  \sum_{x \in N_8(x)} \mathbb{I}(x \in \bs{b})=2,
\end{equation}
where \(N_8(x)\) represents the 8-neighborhood~\cite{moore1962machine} of the point \(x\) in \eqref{eq.branchdef} to iteratively connect a branch from one endpoint to another. All endpoints form the endpoint set \(V_0\). Additionally, endpoints within a radius \(a\) are merged into a single endpoint, and the corresponding branches are updated to reflect the merged endpoint. This process constructs a simplified endpoint set ${\bs{V}}_m$ and the set of branches \(\mathcal{B} = \{ \bs{b} \} \supset \mathcal{C}\).

For each branch, we minimize the number of inserted points along the path between two endpoints, ensuring that the dilation of this path with radius \(a\) sufficiently covers the entire branch. Specifically, for two endpoints \(n_s\) and \(n_e\), we insert the minimal number of points \(n_1, \ldots, n_m\) such that:

\begin{equation}\label{eq.pathcover}
  (n_sn_1...n_mn_e) \oplus \mathcal{F} \supset \bs{b}.
\end{equation}
For a path \(n_s n_1 \ldots n_m n_e\), the points \(n_1, \ldots, n_m\) are added to \(\bs{V}_m\), and the segments \((n_s, n_1), \ldots, (n_m, n_e)\) form edges. All vertices constitute the vertex set \(\bs{V}_c\), and all edges form the edge set \(\bs{E}_c\). In this way, we construct the crack graph \(\mathbb{G}_c = \{ \bs{V}_c, \bs{E}_c \}\). For illustration, in Fig.~\ref{fig.graph_construction}(a), we have \(\bs{V}_m = \{ n_1, n_2 \}\), \(\mathcal{C} = \mathcal{B} = \{ \bs{b}_1 \}\), \(\bs{V}_c = \{ n_1, n_2, n_3, n_4, n_5 \}\), and \(\bs{E}_c = \{ (n_1, n_2), (n_2, n_3), (n_3, n_4), (n_4, n_5) \}\).

When robots perform linear coverage tasks, the target region $\mathcal{A}_c = \mathop{{\bigcup}}_{e \in \bs{E}_c} e \oplus \mathcal{H}$ is considered sensed. Therefore, $\mathcal{A}_w = \mathcal{W} \backslash \mathcal{A}_c$ is the area for robots under exploring mode have to cover. We apply Morse cellular decomposition on $\mathcal{A}_w$. The Reeb graph \(\mathbb{G}_w\) consists of the vertex set \(\bs{V}_w\), which includes all critical points, and the edge set \(\bs{E}_w\), where each edge represents the connectivity of a cell between two vertices. We define \(\phi: \mathbb{R}^2 \to \bs{E}_w\) as a bijective function that projects a cell onto an edge. The mapping \(\phi\) establishes a connection between the graph interpretation and the topology of the space, and consequently the manifolds. The bijection reflects the relationship between the flows and edges that will be introduced later.

As proven in~\cite{kaiyanTRO2024}, by combining \(\mathbb{G}_c\) and \(\mathbb{G}_w\), an additional edge set \(\overline{\bs{E}}\) is constructed to ensure connectivity and the existence of at least one shortest Euler path, where the ``shortest" is defined as minimizing the total Euclidean distance between connected vertices. This is achieved by first matching each critical point in \(\mathbb{G}_w\) with its nearest vertex in \(\mathbb{G}_c\), then pairing the remaining odd nodes accordingly. Consequently, we obtain \(\mathbb{G}_\text{aug}\). Furthermore, if we merge every match of critical points in \(\overline{\bs{E}}\) into a single point, we obtain \(\mathbb{G}_\text{sim}^m\), that is,
\begin{equation}
    \mathbb{G}_\text{sim} = \mathbb{G}_\text{aug} / (\bs{V}_m, \bs{V}_w).
    \label{eq.contraction}
\end{equation}
 This corresponds to the stitching operations discussed later. For illustration, the pairs \([c_2, n_1]\) and \([c_3, n_5]\) in Fig.~\ref{fig.graph_construction}(a) are contracted, resulting in \(\mathbb{G}_{\text{sim}}\) as shown in Fig.~\ref{fig.graph_construction}(b).

A naive approach to finding optimal trajectories for $\mathcal{R}$ is to enumerate all possible Eulerian tours, generate trajectories for each tour, and select the most optimal one. Unfortunately, this is computationally infeasible due to the enormous number of possible tours. In fact, according to~\cite{kaiyanTRO2024}, if we disregard the additional vertices introduced by~(\ref{eq.pathcover}), then in the induced graph $\mathbb{G}_{\text{sim}}' = \mathbb{G}_{\text{sim}}[\bs{V}_m]$, every vertex has degree at least 4. By applying the ``BEST Theorem"~\cite{van1987circuits}, the number of possible Eulerian tours is at least
\begin{equation}\label{eq.pos_approx} 
\prod_{v \in \bs{V}_m} (\deg(v) - 1)! \geq \prod_{v \in \bs{V}_m} (4 - 1)! = 6^{|\bs{V}_m|}, \end{equation}
which grows exponentially with the size of $\bs{V}_m$.

\subsection{Morse Bound and Morse Boundedness}
\begin{figure}[t!]
    \centering

    \begin{subfigure}[t]{0.24\textwidth}
        \centering
        \includegraphics[trim=0.4cm 0cm 0.1cm 0cm, clip, width=\textwidth]{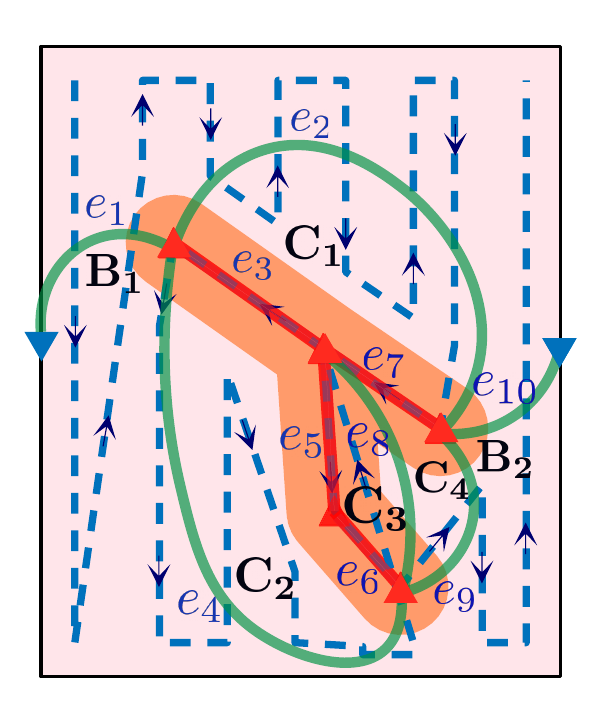}
        \caption{}
    \end{subfigure}
     \hfill
    \begin{subfigure}[t]{0.24\textwidth}
        \centering
        \includegraphics[trim=0.4cm 0cm 0.1cm 0cm, clip, width=\textwidth]{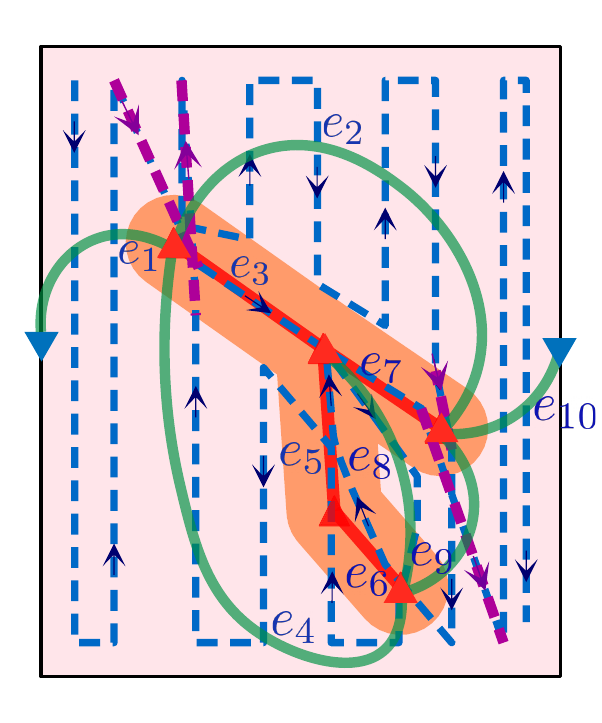}
        \caption{}
    \end{subfigure}
    \caption{Two trajectories for a single-robot double coverage task generated from two Eulerian tours of \(\mathbb{G}_{\text{sim}}\). Blue dotted lines represent the robot trajectories, green lines correspond to Reeb graph edges, and red lines represent linear feature edges. (a) The regulated Eulerian tour ensures continuous trajectory generation across decomposed cells and linear features. (b) The unregulated Eulerian tour results in discontinuous trajectories, with purple dotted lines indicating additional routes required for connectivity.}

    \label{fig:different_Eulerian_tours}

\end{figure}
Not all Eulerian tours need to be considered when searching for an optimal path. In Fig.~\ref{fig:different_Eulerian_tours}, we illustrate two trajectories for a single-robot double coverage task using two different Eulerian tours.
In Fig.~\ref{fig:different_Eulerian_tours}(a), the trajectory is generated using the tour 
$e_1$$\rightarrow$$e_2$$\rightarrow$$e_7$$\rightarrow$$e_3$$\rightarrow$$e_4$$\rightarrow$$e_8$$\rightarrow$$e_5$$\rightarrow$$e_6$$\rightarrow$$e_9$$\rightarrow$$e_{10}\). 
This tour ensures continuity, with no additional connections required to link trajectories within cells or linear features. Furthermore, the trajectory avoids overlapping during traversal.
In contrast, Fig.~\ref{fig:different_Eulerian_tours}(b) shows a trajectory generated by the tour 
\(e_1$$\rightarrow$$e_3$$\rightarrow$$e_8$$\rightarrow$$e_4$$\rightarrow$$e_2$$\rightarrow$$e_9$$\rightarrow$$e_6$$\rightarrow$$e_5$$\rightarrow$$e_7$$\rightarrow$$e_{10}\). 
Discontinuities are observed at $e_1$$\rightarrow$$e_3$, $e_4$$\rightarrow$$e_2$, $e_2$$\rightarrow$$e_9$, and $e_7$$\rightarrow$$e_{10}$, requiring additional connections, as indicated by purple dotted lines. These discontinuities also result in overlapping with previously visited positions and unevenly spaced Boustrophedon trajectories. If this tour were partitioned across multiple robots, conflicts between trajectories are likely to occur at these discontinuities.

\begin{figure}[h!]
    \centering
        \includegraphics[trim=0cm 16.5cm 0cm 0cm, clip, width=0.5\textwidth]{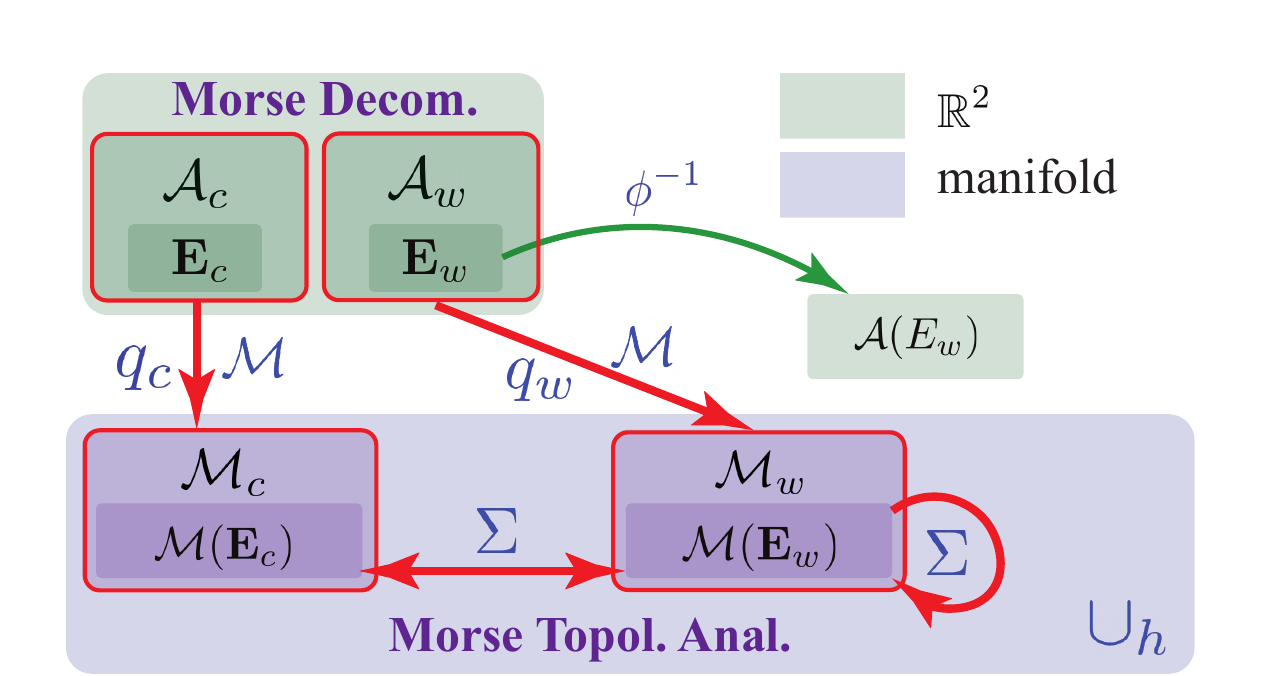}
        \caption{The relationship between Morse cell decomposition in Eucidean space $\mathbb{R}^2$ and the Morse topological analysis in manifold space. $q_c$ and $q_w$ map $\mathcal{A}_c$ and $\mathcal{A}_w$ to $\mathcal{M}_c$ and $\mathcal{M}_w$, respectively. The edge sets ${\bs{E}}_c$ and ${\bs{E}}_w$ both have bijections $\mathcal{M}(\cdot)$ to their manifolds. Across these manifold cells, we further define two quotient maps: stitching ($\Sigma$) and attaching ($\cup_h$).} 
        \label{fig:morse_analysis_framework}
    \label{fig:analysis_framework}
\end{figure}

We aim to eliminate these discontinuities. From a manifold topology aspect, discontinuities may increase or enclose boundaries, or increase holes, which all increase the topological complexity. However, it is difficult to identify this topological complexity in Euclidean space. In this subsection, to quantify the topological complexity 
of each edge connection, we move beyond the conventional analysis approach of applying Morse theory in Euclidean 
space $\mathbb{R}^2$~\cite{howieIJRR2002}. Instead, we transform the problem into 
the Morse theory on a two-dimensional manifold via quotient maps. This approach differentiates 
holes from boundaries and clarifies the number of boundaries in each attaching process. Moreover, 
we introduce a \emph{stitching operation} that captures our task transitions in the manifold context, 
and analyze how stitching affects the attaching maps between manifold cells. Our topological analysis framework is shown in Fig.~\ref{fig:analysis_framework}. Within this framework, discontinuities from \emph{unbounded behavior} emerges once manifold pieces are attached corresponding to individual cells. We then show that any Eulerian tour 
yielding a continuous route must stay \emph{Morse bounded}.

\begin{figure*}[t!]
    \centering
    \begin{subfigure}[t]{0.24\textwidth}
        \centering
        \includegraphics[trim=0.4cm 0cm 0.1cm 0cm, clip, width=\textwidth]{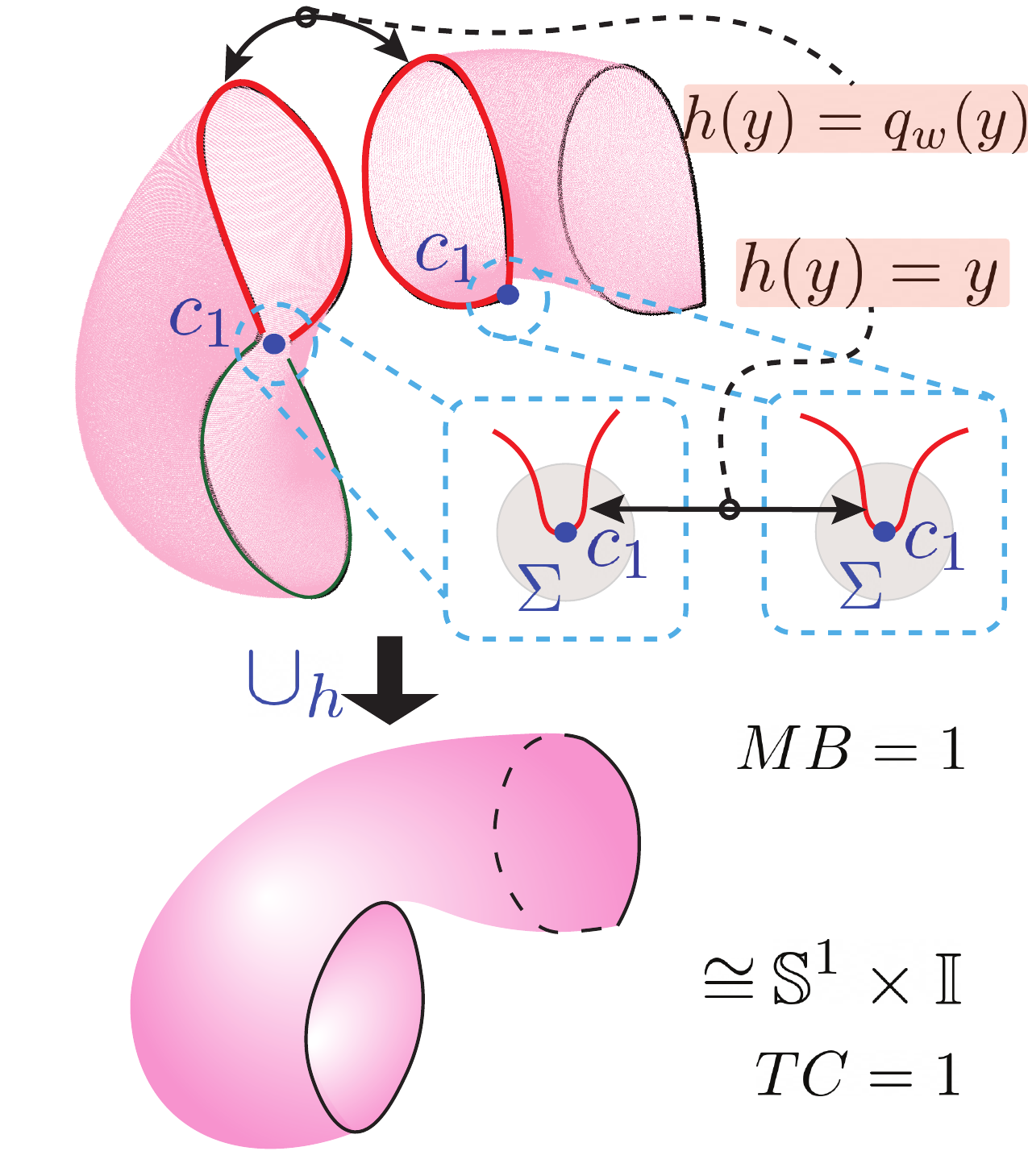}
        \caption{}
    \end{subfigure}
     \hfill
     \vrule height 5cm
    \begin{subfigure}[t]{0.24\textwidth}
        \centering
        \includegraphics[trim=0.0cm 0cm 0.3cm 0cm, clip, width=\textwidth]{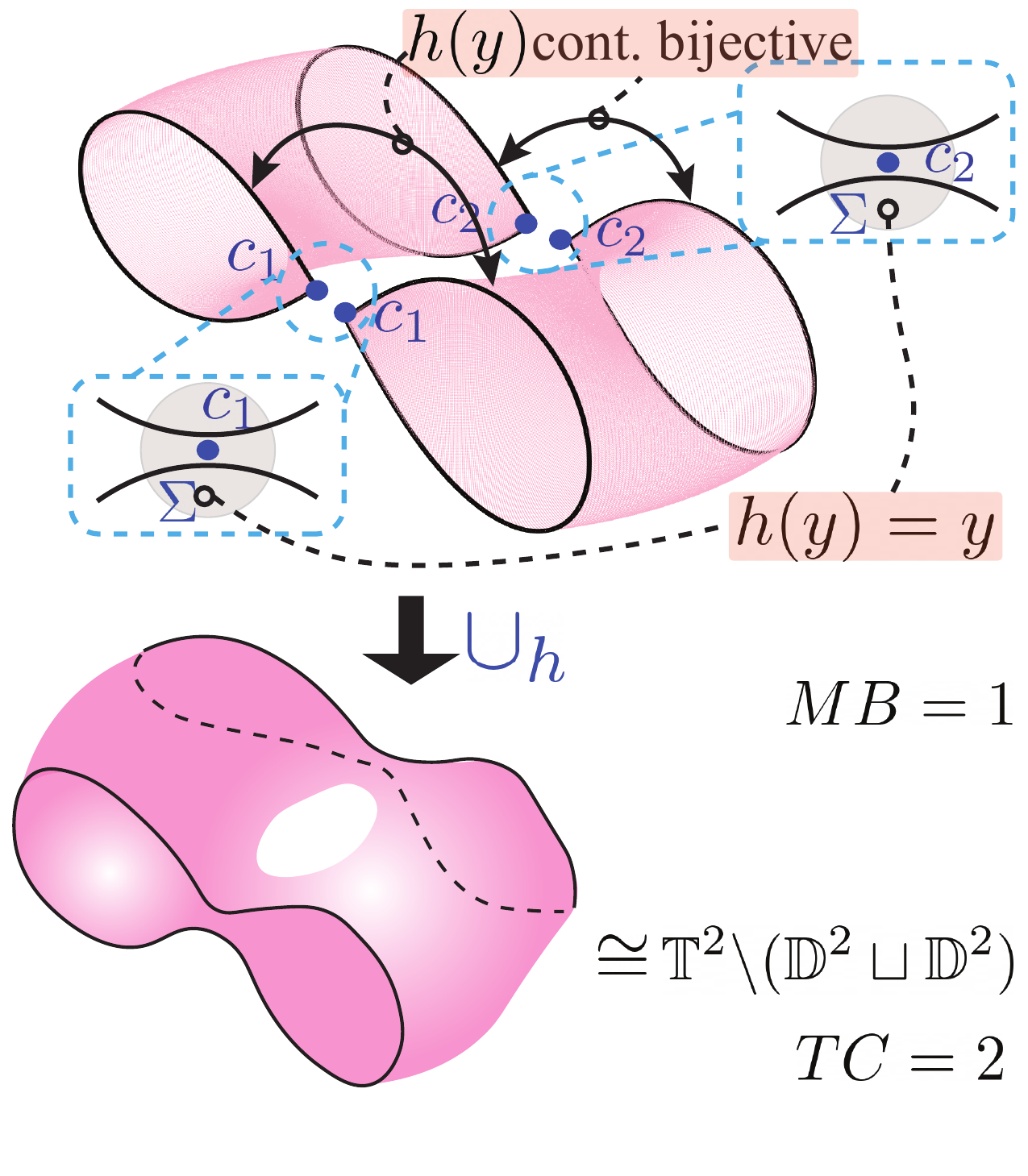}
        \caption{}
    \end{subfigure}
     \hfill
     \vrule height 5cm
    \begin{subfigure}[t]{0.24\textwidth}
        \centering
        \includegraphics[trim=0.4cm 0cm 0.1cm 0cm, clip, width=\textwidth]{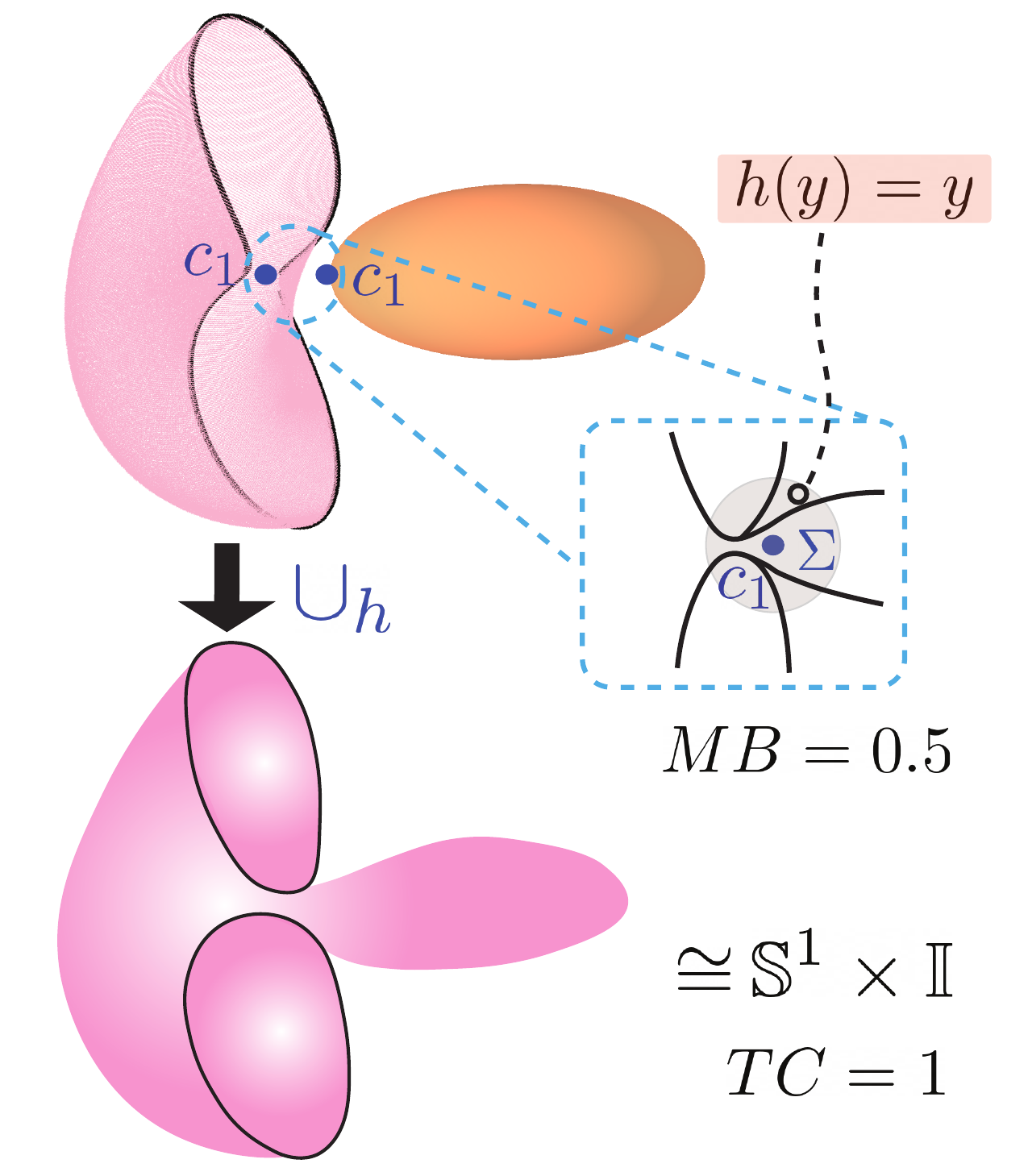}
        \caption{}
    \end{subfigure}
     \hfill
     \vrule height 5cm
    \begin{subfigure}[t]{0.24\textwidth}
        \centering
        \includegraphics[trim=0.4cm 0cm 0.1cm 0cm, clip, width=\textwidth]{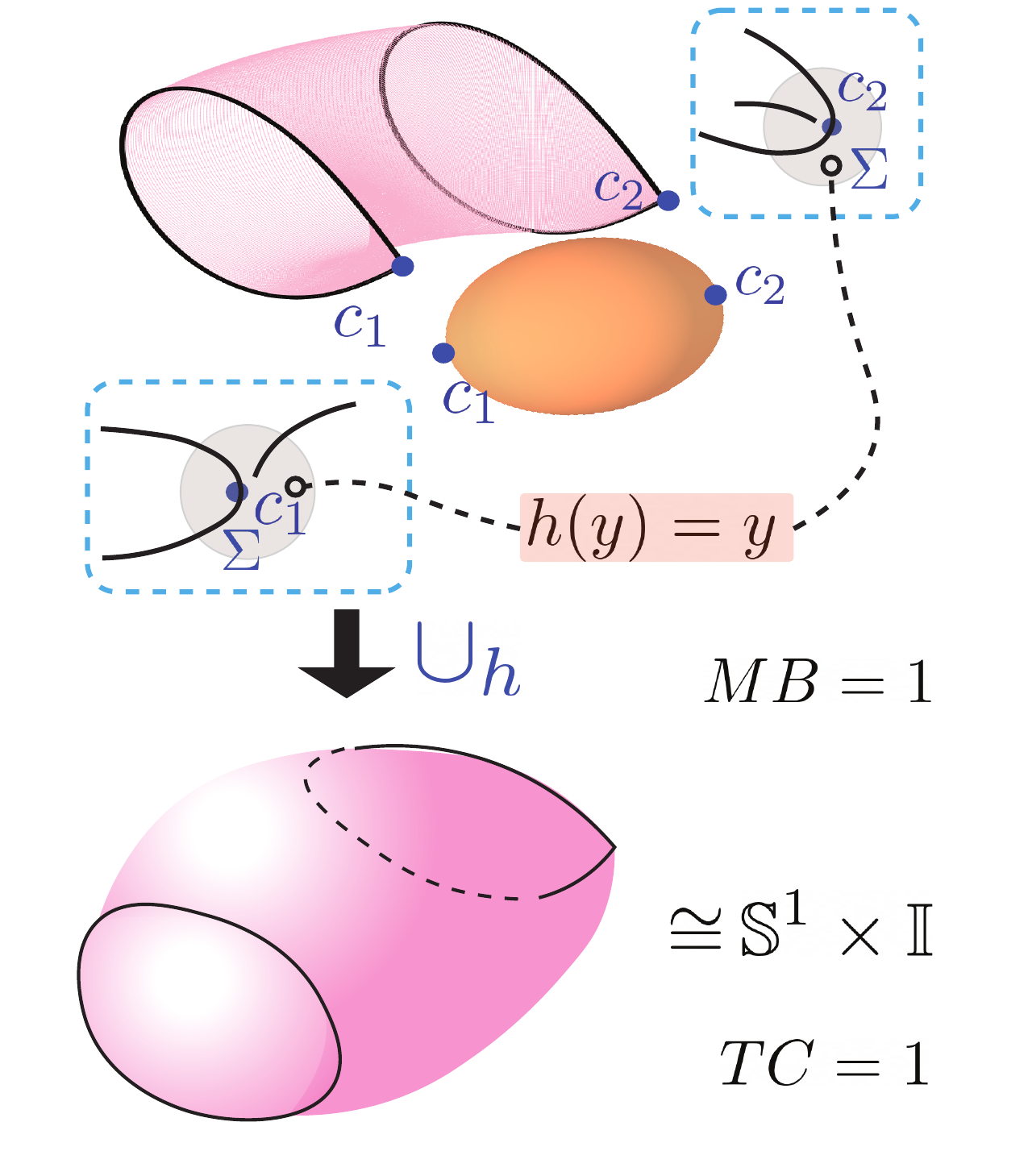}
        \caption{}
    \end{subfigure}
    \caption{Attaching maps of two manifold cells classified into four cases. (a) Two $\mathcal{M}_w$ manifold cells with an attaching relationship defined by $\mathbb{R}^2$ Morse decomposition. (b) Two $\mathcal{M}_w$ manifold cells without an $\mathbb{R}^2$ Morse attachment. (c) and (d) One $\mathcal{M}_w$ manifold cell and one $\mathcal{M}_c$ manifold cell.}

    \label{fig:stitching_process}
\end{figure*}

For the Morse decomposition in $\mathbb{R}^2$, two relationships are maintained: (1) simple connectivity within each cell (strictly 2-cell), and (2) smooth (continuous a.e.) attachability of the cells at each critical point. To transfer this structure into a manifold, we introduce the quotient map
$
q_w: \mathcal{A}_w \to \mathcal{A}_w / \sim.
$
The quotient space \( \mathcal{A}_w / \sim \) is defined by contracting equivalent points, where the equivalence relation is given by the quotient map \( q_w \).
Let $\{\Gamma_\alpha\} $ be cell boundary arcs. If $\Gamma_\alpha$ and $\Gamma_\beta$ are to be glued, we choose an {a.e. continuous} function $f_{\alpha\beta} : \Gamma_\alpha \to \Gamma_\beta$. Then, for $y,z \in \mathcal{A}_w$, we define
\begin{equation}
  y \sim z \quad \overset{\textit{iff.}}{\Longleftrightarrow} \quad
  \begin{cases}
    y = z, 
    & \text{if } y,z \in \mathrm{int}(\mathcal{A}_w), \\[4pt]
    z = f_{\alpha\beta}(y),
    & \text{if } y \in \Gamma_\alpha,\; z \in \Gamma_\beta.
  \end{cases}
  \label{eq:equiv_cartesian_to_manifold}
\end{equation}

Critical points are handled such that saddle-type boundary arcs become loops, whereas minima and maxima collapse to single points. $q_w$ is constructed this way because saddle points support multiple independent flows due to their mixed stable and unstable directions, requiring spatially curved loops to accommodate them, whereas minima and maxima allow only a single independent flow. Under these identifications, 
the quotient space $\mathcal{M}_c = \mathcal{A}_w / \!\sim$ forms a manifold 
whose cells mirror the original Morse decomposition of $\mathcal{A}_w$. 
To distinguish them, we call these entities \emph{manifold cells}. 
Because $\phi$ is a bijection, we write $\mathcal{M}(e_i)$ to denote the 
manifold cell corresponding to the region for the edge $e_i$. In Fig.~\ref{fig.graph_construction}(c), we illustrate the process of constructing $\mathcal{M}(e_1)$ and $\mathcal{M}(e_2)$. $\Gamma_{\alpha}$ and $\Gamma_\beta$ have attachment by $\mathbb{R}^2$ Morse decomposition. Therefore $f_{\alpha \beta}: \Gamma_\alpha \rightarrow \Gamma_\beta$ is defined. $y_4 = f_{\alpha \beta}(y_3)$, so in the quotient space $\mathcal{M}_w$, $q_w(y_3)$ and $q_w(y_4)$ are contracted. In the interior, $y_1 \neq y_2$ leads to $y_1 \nsim y_2$.

We also transform $\mathcal{A}_c$ into a manifold $\mathcal{M}_c$ via a quotient map $q_c$. 
Alternatively, each connected component of $\mathcal{C}$ is topologically a sphere $\mathbb{S}^2$. 
Since $\mathcal{A}_c$ is the complement of $\mathcal{A}_w$, the boundary of $\mathcal{M}_c$ matches the boundary 
of the holes of $\mathcal{M}_w$. We define this boundary as,
\begin{equation}
\label{eq:manifold_boundary}
    \partial\mathcal{M}_c \cap \partial\mathcal{M}_w=\partial\mathcal{A}_c \cap \partial\mathcal{A}_w.
\end{equation}
Since, initially, the whole manifold is $\mathcal{M} = \mathcal{M}_c \sqcup \mathcal{M}_w$, its boundary is given by (\ref{eq:manifold_boundary}) as well. Each edge uses $\oplus$ to project to its area, which is generally not a bijection. However, endpoints are simplified and combined with the line of sight simplification in~(\ref{eq.pathcover}), and thus, we can treat the function as bijective. Consequently, for $ e_i \in \bs{E}_c$, we also use $\mathcal{M}(e_i)$ to represent the corresponding manifold cell. Combined with $\mathcal{M}({\bs{E}}_w)$, every edge in $\mathbb{G}_{\text{sim}}$ has a bijection to its manifold cell, as illustrated in Fig.~\ref{fig.graph_construction} (d). 

In MDC, robots may switch between area exploration and linear-feature service, 
which corresponds to a \emph{stitching} operation, denoted as $\Sigma$, at the task-transition position on $\partial(\mathcal{M})$. This stitching locally joins $\mathcal{M}_c$ and $\mathcal{M}_w$ along their shared boundary, 
thereby enabling a seamless transition between the two coverage modes. Formally, we define the equivalence between $x_c \in \mathcal{M}_c$
and $x_w \in \mathcal{M}_w$ if they lie on the same open ball set of radius $a$ centered at the stitching point and their coordinates match. $\mathcal{M}$ is updated by $\mathcal{M}/ \Sigma$ at each stitching operation.

If stitching is performed \emph{pointwise} along $\partial(\mathcal{M})$, i.e., transitioning tasks continuously, then the boundary defined by~(\ref{eq:manifold_boundary}) would be entirely contracted 
and thus disappear, making $\mathcal{M}$ topologically simply a sphere. However, this is highly inefficient 
since full servicing coverage would then require the robots to perform both coverage tasks solely on $\mathcal{F}$. (\ref{eq.contraction}) ensures that stitching occurs only at critical points, thus 
avoiding continual, unnecessary task transitions. Consequently, we assume 
stitching at every critical point and proceed to discuss the attaching process.

We define \textit{attaching map}, denoted by $\cup_h$, to attach two manifold cells on certain boundaries. An attaching map is a homeomorphism \( h: \partial\mathcal{M}_1 \to \partial \mathcal{M}_2 \) between two manifold cells \( \mathcal{M}_1, \mathcal{M}_2 \) with boundary such that the quotient space \( \mathcal{M}_1 \sqcup \mathcal{M}_2 / h \) forms a new 2-manifold.

By the equivalence defined to construct $\mathcal{M}_w$, we have:
\begin{thm}\normalfont 
    Given two manifold cells (their corresponding edges belong to ${\bs{E}}_w$) whose cells in \( \mathbb{R}^2 \) are attached through the Morse decomposition, there exists an attaching map.
\end{thm}

\begin{pf}
For each pair of two manifold cells, $q_w $ induced by $f_{\alpha \beta}$ is an a.e. continuous bijection on the boundary, and the discontinuity occurs on critical points because the boundary of one manifold does not have the critical points while the other one has. A stitching is defined at each critical point. Stitching is a quotient map on the open balls centered at each critical point. Although one manifold does not contain the critical point on its boundary, this discontinuity is removable, as the remaining boundary within the open ball projects identically onto the boundary of the other manifold. Inside each small open ball, let $h$ be the identity map, which is continuous and bijective. Outside these open balls, let $h$ be $q_w$. The arbitrariness of $f_{\alpha \beta}$ and $q_w$ makes sure the continuity of $h$ on the boundary of the open balls can be satisfied. Then $h$ is a globally continuous bijection, which is a homeomorphism.
\end{pf}

One example of this stitching process and the construction of the homeomorphism is illustrated in Fig.~\ref{fig:stitching_process}(a). 
For two manifold cells which both belong to $\mathcal{M}_c$ and their boundaries intersect, we attach two spheres with an empty continuous function $h: \emptyset \to \emptyset$.

Next we define attaching maps in the remaining two cases combining stitching operations:

(1) \emph{Two Reeb-edge manifold cells whose boundaries intersect, 
  but the attachment is not prescribed by the Morse decomposition.} 
  In this case, the boundaries intersect at one or two critical points only.
  Inside the open ball centered at each critical point, we let $h$ 
  act as the identity map. Outside of these balls, a continuous bijection 
  between the remaining boundary arcs can always be constructed, since each boundary 
  is topologically that of a disk $\mathbb{D}^2$. $h$ is again globally continuous and bijective. An illustration of such attaching map is shown in Fig.~\ref{fig:stitching_process}(b).

(2) \emph{Transition between a Reeb-edge cell and a linear-feature-edge cell.} 
  In this case, $h$ is an identity function inside the small open ball centered at each critical point and on the shared boundary defined exactly by stitching operations. Fig.~\ref{fig:stitching_process}(c) and (d) illustrate this kind of attaching maps.

The purpose of considering the problem in the manifold space is to quantify the topological complexities of the attached manifold at each step. In our problem, a hole inside a manifold would directly cause overlapping in the Boustrophedon trajectory generation, while a boundary would cause discontinuous connections at worst. In the $\mathbb{R}^2$ Morse topology, it is difficult to identify holes and the number of boundaries. For a 2-manifold $\mathcal{M}^\text{att}$, we use $N_\text{g}$ to represent the genus, and $N_\text{bd}$ to represent the number of boundaries, then the topological complexity is calculated as,

\begin{equation}
    TC(\mathcal{M}^\text{att}) = N_\text{g}(\mathcal{M}^\text{att})+\frac{1}{2}N_{\text{bd}}(\mathcal{M}^\text{att}).
    \label{eq:topological_complexity}
\end{equation}

For example, in Fig.~\ref{fig:stitching_process}(b), the attached manifold has two boundaries and one hole, therefore $TC=1+\frac{1}{2}\times2 = 2$. (\ref{eq:topological_complexity}) measures the level of discontinuity for the trajectory generation within this manifold.

Within an Eulerian tour, we wish to avoid any discontinuity in trajectory generation. 
In topological terms, there should be no discontinuous manifold attachments along the tour. 
More precisely, within any sub-path of the Eulerian tour, the topological complexity of the attached 
manifold must not exceed that of the most complex manifold cell involved in that sub-path. For an Eulerian tour $p$ and a sub-path $p'$, we define the Morse bound, denoted as ${MB}$, as the following:
\begin{equation}
    MB(p') = \max \left\{TC(\mathcal{M}(e)) \mid e \in p' \right\},
\end{equation}
and the attached manifold complexity
\begin{equation}
    TC(\mathcal{M}^{p'}) = TC(\mathop{\cup_h}_{e \in p'}(\mathcal{M}(e)).
\end{equation}
\noindent
We summarize the trajectory continuity condition as follows:
\begin{corollary}\label{thm:continuous_morse_bound}
    An Eulerian tour $p$ that produces a continuous route is \textbf{Morse bounded} from 
    start to finish, namely, for any $p' \subseteq p$, 
    $TC(\mathcal{M}^{p'}) \;\leq\; MB(p')$.\end{corollary}
As an example, attaching maps in Figs.~\ref{fig:stitching_process}(a) and (d) hold Morse boundedness, but those in Figs.~\ref{fig:stitching_process}(b) and (c) do not. To conclude, we quantify the manifold's topological complexity at each attaching map 
and use  {Morse bounds} to regulate the behavior of an Eulerian tour, ensuring 
continuity throughout.

\subsection{Extension to Multiple Robots}
When extending to the multi-robot case, Eulerian tours that are equivalent in terms of Morse invariance are not necessarily equivalent for multi-robot allocation due to the differenet weights among each edge. In fact, deploying the FHK algorithm~\cite{Xu-2011-7344} to solve the $k$-RPP problem generally provides a solution that is at most a ($2-\frac{1}{N}$) approximation of the optimal allocation. This limitation arises because the Eulerian tour is selected randomly, which may not align with the optimal distribution of tasks among multiple robots.

To find an optimal solution for multi-robot allocation, it is necessary to span all possible Eulerian tours that validly attach manifold cells from the first edge to the last. This corresponds to identifying the Morse-bounded collection:
\begin{equation}\label{eq.equi_class}
\mathcal{P} = 
\left\{
p \mid p \text{ is Eulerian and Morse bounded}
\right\}.
\end{equation}
The allocation solution is then determined by searching for the optimal solution within this collection. By focusing on this subset of Eulerian tours, the approach ensures that the solution is both Morse-bounded and optimized for multi-robot task allocation.

\section{Hierarchical Cyclic Merging Regulation}

In this section, we establish a connection between Morse boundedness and specific regulation rules for constructing Eulerian tours on $\mathbb{G}_{\text{sim}}$. Instead of blindly searching for Eulerian tours, we employ HCMR, a heuristic algorithm derived directly from these regulation rules. 

\subsection{Regulation Rules on Graph}

Morse boundedness is a property of the manifold space. To implement specific algorithms, regulation rules must be established on the graph. The bijection between the manifold cells and the edge set of $\mathbb{G}_{\text{sim}}$
  ensures the feasibility of defining such rules.

A closed loop in $\mathbb{G}_{\text{sim}}$
  implies that the attaching process in the manifold space returns to the initial manifold cell upon completing the loop traversal. Therefore, within a loop, commutativity of attachments holds, which implies traversal order can be switched without affecting the final topology. However, a traversal within a loop does not necessarily guarantee Morse boundedness.  In Fig.~\ref{fig:stitching_process}(d), $e_2$ and $e_5$ form a loop, and the attached manifold is bounded by the Morse bound. In contrast, in Fig.~\ref{fig:stitching_process}(b), $e_2$ and $e_3$ also form a loop, but Morse boundedness is violated.

\begin{figure}[h!]
    \centering
    \begin{subfigure}[t]{0.235\textwidth}
        \centering
        \includegraphics[trim=1.4cm 1cm 0cm 2cm, clip, width=\textwidth]{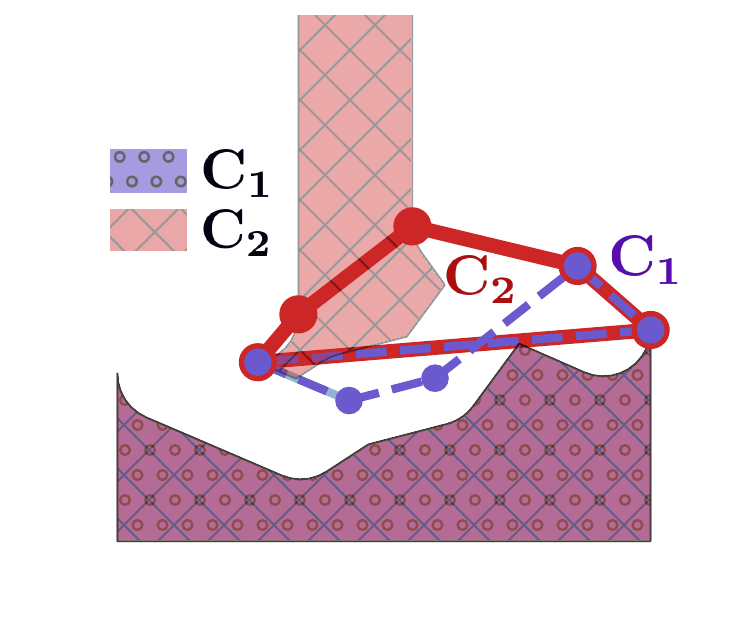}
        \caption{}
    \end{subfigure}
    \hfill
    \begin{subfigure}[t]{0.235\textwidth}
        \centering
        \includegraphics[trim=1.6cm 0.7cm 1.3cm 2cm, clip, width=\textwidth]{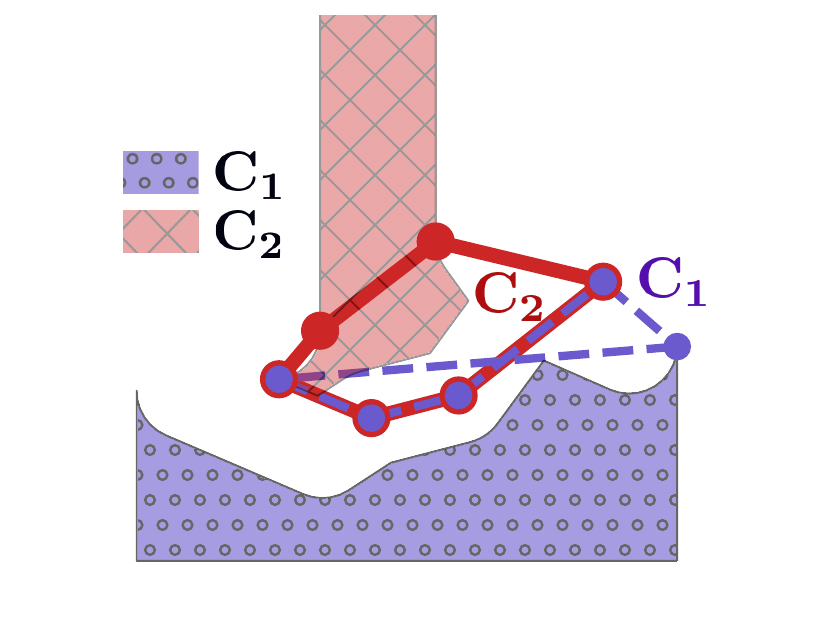}
        \caption{}
    \end{subfigure}
    \caption{Illustration of the definition of the disjoint cycle basis \(\mathcal{K}^d\). (a) The union of Reeb cells of cycle \(C_2\) completely contains (or covers) those of \(C_1\), so \(C_1\) and \(C_2\) cannot both belong to a disjoint cycle basis. (b) \(C_2\) does not cover \(C_1\) in the Reeb sense, so \(C_1\) and \(C_2\) can both belong to a \(\mathcal{K}^d\).}

    \label{fig:cycle_basis}
\end{figure}

In the context of algorithm deployment, parallel edges form loops but coincide, so in the following discussion, we consider only cycles. We define a \emph{disjoint cycle basis} $\mathcal{K}^d$ 
as follows: a cycle basis $\mathcal{K}^d$ is \emph{disjoint} if, for any two distinct 
cycles $C_i, C_j \in \mathcal{K}^d$,
\begin{equation}
\label{eq.disjoint_cycle_basis}
    \bigcup_{e \in C_i \cap \mathbf{E}_w} \phi^{-1}(e)
    \;\nsubseteq\;
    \bigcup_{e' \in C_j \cap \mathbf{E}_w} \phi^{-1}(e').
\end{equation}
That is,
no cycle in $\mathcal{K}^d$
  has its Reeb cells fully contained by those of any other cycle in the basis. Fig.~\ref{fig:cycle_basis} illustrates the concept. Note that this disjoint 
cycle basis need not be unique, but finding one suffices for our purposes.

\begin{lemma}
  A cycle preserves Morse boundedness if and only if that cycle belongs to a disjoint cycle basis.
  \label{lemma:cycle_basis_morse_bounded}
\end{lemma}

\begin{pf}
(\(\Rightarrow\)) Suppose, for contradiction, that there exist cycles \(C\) and \(C'\) such that \(C\) fully contains all the Reeb cells of \(C'\). Then in the merged manifold already containing \(\mathcal{M}(C)\), the manifold cell \(\mathcal{M}(C')\) has two distinct boundary components (its two “ends”) that must be stitched to corresponding portions of \(\partial\mathcal{M}(C)\). However, the “outside” portion of \(\partial\mathcal{M}(C')\) cannot consist entirely of Reeb edges; if it did, there would be no critical point to separate it from the “inside” boundary, and the two boundaries would be merely parallel—that is, \(C\) would become exactly \(C'\) in the multi-graph setting. Since at least one edge from \(\mathbf{E}_c\) is present, the Minkowski sum forces the duplicate boundary (see Fig.~\ref{fig:stitching_process}(b)) to persist upon stitching, thereby introducing an extra hole in the merged manifold and violating Morse boundedness.

(\(\Leftarrow\)) Conversely, assume that cycle \(C\) belongs to a disjoint cycle basis. Then its associated manifold cell \(\mathcal{M}(C)\) is not fully enclosed by any other, so its boundary \(\partial\mathcal{M}(C)\) appears only once in any merged manifold. Hence, when \(\mathcal{M}(C)\) is stitched and attached, each portion of \(\partial\mathcal{M}(C)\) is paired one-to-one with the corresponding boundary segment, and no duplicate boundary (and thus no extra hole) arises. This completes the proof.

 \end{pf}

\begin{figure*}[h!]
    \centering
    \includegraphics[trim=0cm 0cm 2cm 0cm, clip, width=\textwidth]{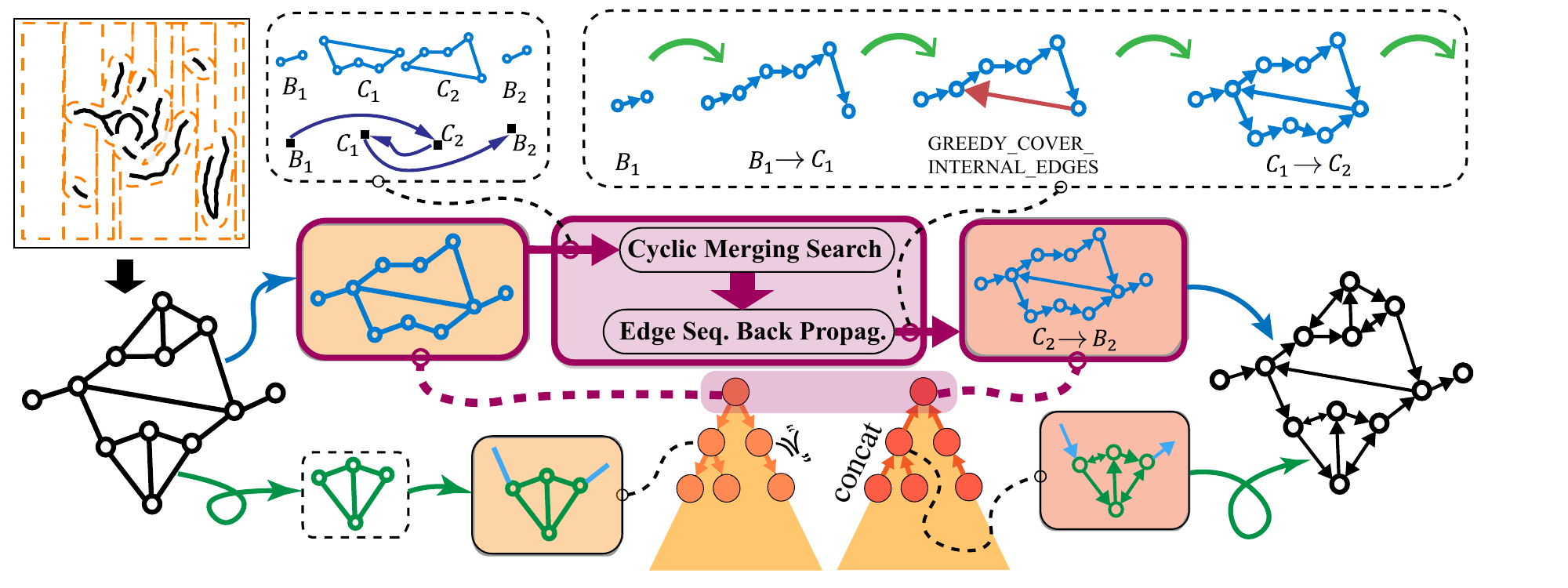}
    \caption{The overview of HCMR framework. The simplified combined graph $\mathbb{G}_\text{sim}$ is first decomposed into different levels of hierarchy. On each component in each level, a Cyclic Merging Search (CMS) is performed to regulate the order of edge sequences. Then Edge Back Propagation is performed for the generation of the Eulerian tour of the component. Edge sequences are concatenated iteratively through the hierarchy to obtain the collection of Eulerian sequences that are Morse bounded.}
    \label{fig:pipeline}
\end{figure*}

$\mathcal{K}^d$ spans the cycle space ${\bs{E}}_{\text{cycle}}$, and along with the bridges, they collectively cover all edges in $\mathbb{G}_{\text{sim}}$. Viewing each cycle or bridge as a single unit, the objective is to find a valid sequence of these units whose traversal spans all edges while preserving Morse boundedness throughout.

\begin{lemma}
    Two cycles $C_1, C_2$ (not necessarily in $\mathcal{K}^d$) preserve Morse boundedness upon merging 
    if and only if their intersection is exactly one simple path.
    \label{lemma:cycle_seq}
\end{lemma}

\begin{pf}
Gluing $\mathcal{M}(C_1)$ and $\mathcal{M}(C_2)$ along a single shared boundary path does not form extra boundaries or holes. If the intersection is a point, the attaching map at this point breaks a boundary into two. Multiple paths indicate extra holes.
\end{pf}

\begin{lemma}
    A bridge $B$ merging with a cycle edge $e_C$ preserves Morse boundedness 
    if and only if $e_C \in \mathbf{E}_w$.
    \label{lemma:bridge_to_cycle}
\end{lemma}

\begin{pf}
 {(\(\Rightarrow\))} If $e_C \in \mathbf{E}_w$, notice that $B$ is a bridge, then the attaching map is induced by $q_w$. The attaching relationship defined by $\mathbb{R}^2$ Morse decomposition considering the minimum spanning tree preserves Morse boundedness.

 {(\(\Leftarrow\))} If $e_C \in \mathbf{E}_c$, the attaching map breaks a boundary into two separate boundaries.
\end{pf}

\subsection{Hierarchy Construction}
We define a \emph{contain} relation $\succcurlyeq$ among cycles in 
$\mathcal{K}^d$ as follows. Let
$
  g_{\min}(C) \;=\; \min\{\,g(n)\mid n\in C\}$, $ 
  g_{\max}(C) \;=\; \max\{\,g(n)\mid n\in C\}.
$
Then, for any $C_i, C_j \in \mathcal{K}^d$, we write $C_i \succcurlyeq C_j$ if
$ \bigl[g_{\min}(C_i),\, g_{\max}(C_i)\bigr]
  \;\subset\;
  \bigl[g_{\min}(C_j),\, g_{\max}(C_j)\bigr].
$
By definition, a cycle can never contain itself. Similarly, $\succcurlyeq$ can be extended to describe the relationship between a connected component and a cycle.



We use this relationship to construct the hierarchy for cycles as: If all edge-adjacent cycles contain a cycle \(C_i\) , \(C_i\) is ``local" relative to them, as its \(g(p)\) interval is fully contained. Thus, \(C_i\) belongs to the next hierarchy level as their child. Hierarchy construction continues iteratively until no further changes occur. 

For hierarchy level \( i \), we first generate an induced subgraph \(\mathbb{G}_{\text{ind}}\) containing all vertices in the cycles at level \( i \). Next, bridges in the bridge space of \(\mathbb{G}_\text{sim}\) that share vertices with \(\mathbb{G}_{\text{ind}}\) are added to form \(\mathbb{G}_i\). We decompose \(\mathbb{G}_i\) into connected components, and for every component $\mathbb{G}_{i,j}$,  by the construction of \(\mathbb{G}_\text{sim}\) we have:

\begin{lemma}\label{rmk.contain_conn}
  Suppose there is a cycle $C_n \in \mathcal{K}^d$ such that 
  $\mathbb{G}_{i,j} \preccurlyeq C_n$. Then exactly one edge in $\bs{E}_w$ 
  has the largest $x$-coordinate equal to 
  $ g_{\min}(\mathbb{G}_{i,j})$, 
  and exactly one edge in $\bs{E}_w$ 
  has the smallest $x$-coordinate equal to 
  $g_{\max}(\mathbb{G}_{i,j})$.
\end{lemma}
\begin{pf}
Since $\mathbb{G}_{i,j} \preccurlyeq C_n$ and $C_n \in \mathcal{K}^d$, 
we know $g(p)$ for all $p\in \mathbb{G}_{i,j}$ lies strictly within 
the $g$-range spanned by $C_n$. In the Reeb decomposition, each cycle in 
$\mathcal{K}^d$ contains exactly one ``leftmost'' and one ``rightmost'' 
Reeb edge (otherwise it could not form a single closed loop). Thus, if there 
were two or more Reeb edges in $\bs{E}_w$ sharing the same minimal (or maximal) 
$g$-value for $\mathbb{G}_{i,j}$, it would contradict the requirement that 
$C_n$ maintains a unique boundary arc on each side. 
\end{pf}

For each component $\mathbb{G}_{i,j}$, if the leftmost edge(s) is not a bridge, the edge in \( \bs{E}_w \) with the maximum $x=g_{\min}(\mathbb{G}_{i,j})$ is added by Lemma~\ref{rmk.contain_conn}, and similarly with the rightmost edge(s). Notice that these two edges become bridges in the subgraph $\mathbb{G}_{i,j}$.
\subsection{Cyclic Merging Search}






\begin{algorithm}
\DontPrintSemicolon
\caption{Cyclic Merging Search (CMS)}\label{algo:legalpaths}
\SetAlgoVlined
\SetKwInOut{Input}{Input}
\SetKwInOut{Output}{Output}
\Input{${\bs{E}}_{\text{cycle}}$, ${\bs{E}}_{\text{bridge}}$, $B_s$, $B_e$}
\Output{$\mathcal{S}$}
$\mathcal{S} \gets \emptyset$, $\mathcal{V}_C \gets \emptyset$, $V_B \gets \{ B_s \}$, $\bs{m} \gets \emptyset$ \;
\SetKwFunction{FExplore}{Explore} 
\SetKwProg{Fn}{Function}{:}{\KwRet}

\Fn{\FExplore{$\bs{S}$, $V_C$, $V_B$, $\bs{m} $}}{
    \If{$V_C ={\bs{E}}_{\text{cycle}}$ \textbf{and} $\bs{S}[-1] = b_e$}{
        $\mathcal{S} \gets \mathcal{S} + \bs{S} $\;
    }
    \If{\text{connectivity} \textbf{and not} \text{has\_hole}}{
        \uIf{$s \gets \bs{S} [-1]$ is a cycle}{
            \ForEach{$C \in (\text{Adj}^v(S) \cap \text{Adj}^e(\bs{m})) \setminus V_c$}{
                \FExplore{$\bs{S} + C$, $V_C \cup  C$, $V_b$, $\bs{m} \triangle  C$}\;
            }
            \If{$(\text{Adj}^v(S) \cap \text{Adj}^e(\bs{m})) \setminus V_C=\emptyset$}{
                \ForEach{$B \in \text{Adj}^v(S) \setminus V_B$, $b \ne b_e$}{
                    \FExplore{$\bs{S} + B$, $V_C$, $V_B \cup  B$, $\emptyset$}\;
                }
            }
        }
        \Else{
            \ForEach{$S' \in \text{Adj}^v(S) \setminus (V_C \cup V_B)$}{
                \FExplore{$\bs{S} + S'$, $V_C$, $V_B$, $\emptyset$}\;
            }
        }
    }
}

\end{algorithm}
Cyclic Merging Search (CMS) is performed on each component \(\mathbb{G}_{i,j}\) as described in Algorithm \ref{algo:legalpaths}, which is a depth-first search algorithm with pruning. In this and the following subsection, a segment \(s\) can be a bridge or a cycle, i.e., \(s \in \{\text{bridge}, \text{cycle}\}\). Using Lemma~\ref{lemma:cycle_seq}, we propose the following heuristic rule to maintain Morse boundedness while preserving the Eulerian property:

(1) \textit{Current is bridge:} If the current segment in the recursive exploration is a bridge, the next segment can be any unvisited vertex-adjacent segment \(S' \in \text{Adj}^v(S) \setminus (V_C \cup V_B)\), where \(V_C\) and \(V_B\) are the sets of visited cycles and bridges, respectively. The algorithm updates the current sequence of segments \(S\) to include \(S'\) and resets the merged cycle \(m\) to \(\emptyset\), preparing for the formation of new cycles (line 14).

(2) \textit{Current is cycle:}  When there are unvisited cycles adjacent to the current cycle, the algorithm selects each cycle \(c'\) that is edge-adjacent to the merged cycle \(m\) and vertex-adjacent to the current cycle \(s\). The former selects candidate cycles for exploration and merging (by Lemma~\ref{lemma:cycle_seq}). However, whether $\bs{m}$ and $S$ intersect with one simple path is not promised until here. The latter guarantees the possibility of forming an Eulerian path by maintaining vertex continuity. The symmetric difference operation \(\bs{m} \triangle  C \) is applied to update the merged cycles, and the exploration continues recursively (lines 7--8). If no adjacent cycles are available, the algorithm explores unvisited bridges connected to the current cycle. This allows the exploration to branch out into new regions of the graph (lines 9--10).

(3) \textit{Connectivity and hole checks:} Before exploring any segment, the algorithm ensures that the graph remains connected and free of holes (line 5). The graph is considered connected if, after removing edges corresponding to visited cycles and bridges, at most one disjoint component exists outside the vertex set of higher-level cycles. Otherwise, the remaining graph has no chance to be spanned, and therefore, the exploration is prunned. In addition, by Lemma \ref{lemma:cycle_seq}, we need to check if $\bs{m} $ and $S$ intersect with multiple paths. This is equivalent to verify that unvisited cycles are not entirely enclosed by visited cycles and bridges. If any unvisited cycle’s edge set is a subset of visited edges, a ``hole" is detected, and exploration stops.

(4) \textit{Recording legal sequences of segments:} Once all cycles and bridges in the graph have been visited and the sequence of segments ends at the specified ending bridge \(b_e\), the current sequence \(S\) is added to the set of legal sequences \(\mathcal{S}\) (lines 3--4).

In CMS, we iteratively apply Lemma \ref{lemma:cycle_seq} at each searching and pruning step, obtaining a collection of segment sequences that serve as regulations throughout the construction of Eulerian tours.

\subsection{Back to Eulerian Tour}

We use each sequence of segments \( \bs{S} \) in \( \mathbb{G}_{i,j} \) to regulate Eulerian tour generation. However, if the segment is a cycle, the \(\kappa\)-transformation indicates that even within this cycle, there are two possible directions (clockwise and counterclockwise) to span it. To determine the direction that best encourages Morse boundedness, we apply a greedy heuristic Edge Sequence Back Propagation (BP) algorithm, as detailed in Algorithm~\ref{algo:edge_seq_backprop}.

\begin{algorithm}
\DontPrintSemicolon
\caption{Edge Sequence Back Propagation (BP)}\label{algo:edge_seq_backprop}
\SetAlgoVlined
\SetKwInOut{Input}{Input}
\SetKwInOut{Output}{Output}
\Input{$\bs{S}$, ${\bs{E}}_{\text{bridge}}$, $\mathbb{G}_{i,j}$}
\Output{$p$}
$v_{\text{curr}} \gets {\bs{E}}_{\text{bridge}}[0][0]$, $\bs{m} \gets \emptyset$\;

\For{$i \gets 0$ \KwTo $|S| - 1$}{
   $S \gets \bs{S} [i]$,$S_{\text{next}} \gets \bs{S} [i + 1]$\;
    
    \uIf{\textsc{IsBridge}($S$)}{
        $\bs{m} \gets \emptyset$, \textsc{BridgeWalk}($v_{\text{curr}}, S$)\;
    }
    \ElseIf{\textsc{IsCycle}($S$)}{
        \uIf{$\bs{m} \neq \emptyset$}{
            \textsc{GreedyCoverInternalEdges}($v_{\text{curr}}, m, S$)\\
            $\bs{m} \gets \bs{m} \triangle S$\;
        }
        \lElse{ $\bs{m} \gets S$}
        \textsc{CycleToNextSegment}($v_{\text{curr}}, S, S_{\text{next}}$)\;
    }
}
\end{algorithm}

(1) \textit{Current segment is a bridge:} If the current segment \(S\) is a bridge, the algorithm first resets the merged cycle \( \bs{m} \) to \( \emptyset \). Then we perform a \textsc{BridgeWalk} starting from the current node \( v_{\text{curr}} \) and also update $v_{\text{curr}}$ (line 5).

(2) \textit{Current segment is a cycle:} If the merged cycle is not $\emptyset$, the algorithm invokes the function \textsc{GreedyCoverInternalEdges} (line 8) to greedily cover the internal edges shared between the merged cycle \( \bs{m} \) and the current cycle \( S \). According to Lemma \ref{lemma:cycle_basis_morse_bounded}, connecting edges within the same disjoint cycle preserves Morse invariance. By greedily covering these internal edges, the algorithm ensures that Morse boundedness is maintained not only within the current cycle but also across the merged cycle, since these internal edges are the common edges shared by \( \bs{m} \) and \( S \) by construction. Morse boundedness is passed by through the shared edges. Furthermore, greedily spanning internal edges ensures that the path covers all edges in the subgraph \( G_{i,j} \), thereby contributing to the completeness of the Eulerian path. After covering the internal edges, the merged cycle \( \bs{m} \) is updated with \( \bs{m} \triangle S \) (line 9).

If the merged cycle is empty, this indicates transitioning from a bridge to a cycle. The algorithm sets \( \bs{m} \) to the current cycle (line 10) to start forming a new merged cycle. Here, the traversal of the cycle begins from an edge in Reeb graph $\bs{E}_w$ connected to the current node \( v_{\text{curr}} \) by Lemma \ref{lemma:bridge_to_cycle}.

Next, the algorithm traverses the remaining unvisited edges $E_{\text{seq}}$ within the current cycle (line~11). For each edge $(u, v) \in E_{\text{seq}}$, it checks whether the next vertex $v$ is part of the vertex set of the next segment. If so, this indicates a transition to the next segment, and the loop breaks. Otherwise, the path continues forward along $E_{\text{seq}}$.

\begin{algorithm}[t]
\DontPrintSemicolon
\caption{Compute Eulerian Tours via Hierarchy Iteration}\label{algo:hierarchy_edge_sequences}
\SetAlgoVlined
\SetKwInOut{Input}{Input}
\SetKwInOut{Output}{Output}
\Input{$H$, $\mathcal{C}$, $\mathbb{G}$}
\Output{$\mathcal{P}$}
$\mathcal{P} \gets \emptyset$\;
\ForEach{level $i$ in hierarchy $H$}{
    \ForEach{connected component $\mathbb{G}_{i,j}$ at level $i$}{
        $\mathcal{S}_{i,j} \gets \textsc{CMS}(\mathbb{G}_{i,j})$, 
        $\mathcal{E}_{i,j} \gets \textsc{BP}(\mathcal{S}_{i,j})$\;
    }
    \uIf{$i = 0$}{
        $\mathcal{P} \gets  \left\{\mathcal{E}_{0,j}|\text{ } \forall j \right\}$\; 
    }
    \lElse{
        $\mathcal{P} \gets \{\text{Concat}(p, e) \mid p \in \mathcal{P}, e \in \prod \mathcal{E}_{i,j}\}$
    }
}
\end{algorithm}

Finally, by integrating the constructed hierarchy and iteratively concatenating paths, as outlined in Algorithm~\ref{algo:hierarchy_edge_sequences}, we obtain the following results.

\begin{thm}\normalfont\label{thm.path_exist_unique}
    Given fixed segment sequences at all hierarchy levels, the concatenated path forms an Eulerian tour for $\mathbb{G}_\text{sim}$.
\end{thm}

\begin{pf}
    Within each component, CMS ensures that all segments are spanned exactly once. In BP, the function \textsc{GreedyCoverInternalEdges} guarantees that every internal edge within a cycle cluster is visited exactly once. Since an Eulerian walk exists, the contour of each cluster is also traversed. By concatenating the Eulerian tours across all hierarchy levels, a single route that visits every edge of $\mathbb{G}_\text{sim}$ exactly once is obtained, thus forming an Eulerian tour.
\end{pf}

By Theorem~\ref{thm.path_exist_unique}, we obtain the Morse bounded collection as stated in~(\ref{eq.equi_class}), and the size of this set is
\begin{equation}\label{eq.equi_class_size}
    |\mathcal{P}|=\mathop{\prod}\limits_{l \in LH} \mathop{\prod}\limits_{c \in CC(\mathbb{G}_{l})} |S_{c}|,
\end{equation}
where $LH$ is the level set of hierarchy $H$, and $CC$ is the function that obtains the connected components for a graph.

\begin{figure*}[h!]
    \centering
    \begin{subfigure}[t]{0.24\textwidth}
        \centering
        \includegraphics[, width=\textwidth]{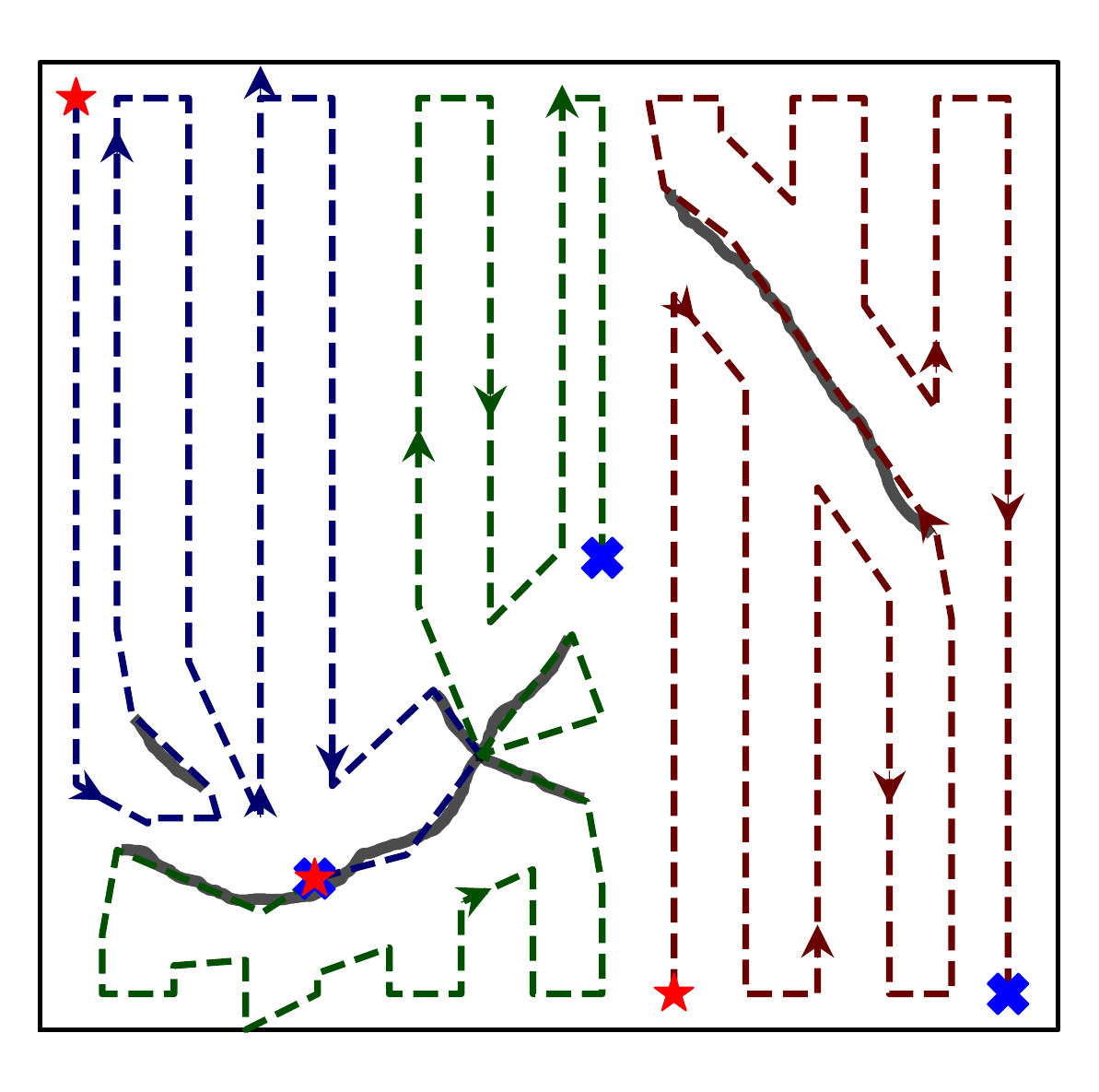}
    \end{subfigure}
    \begin{subfigure}[t]{0.24\textwidth}
        \centering
        \includegraphics[ width=\textwidth]{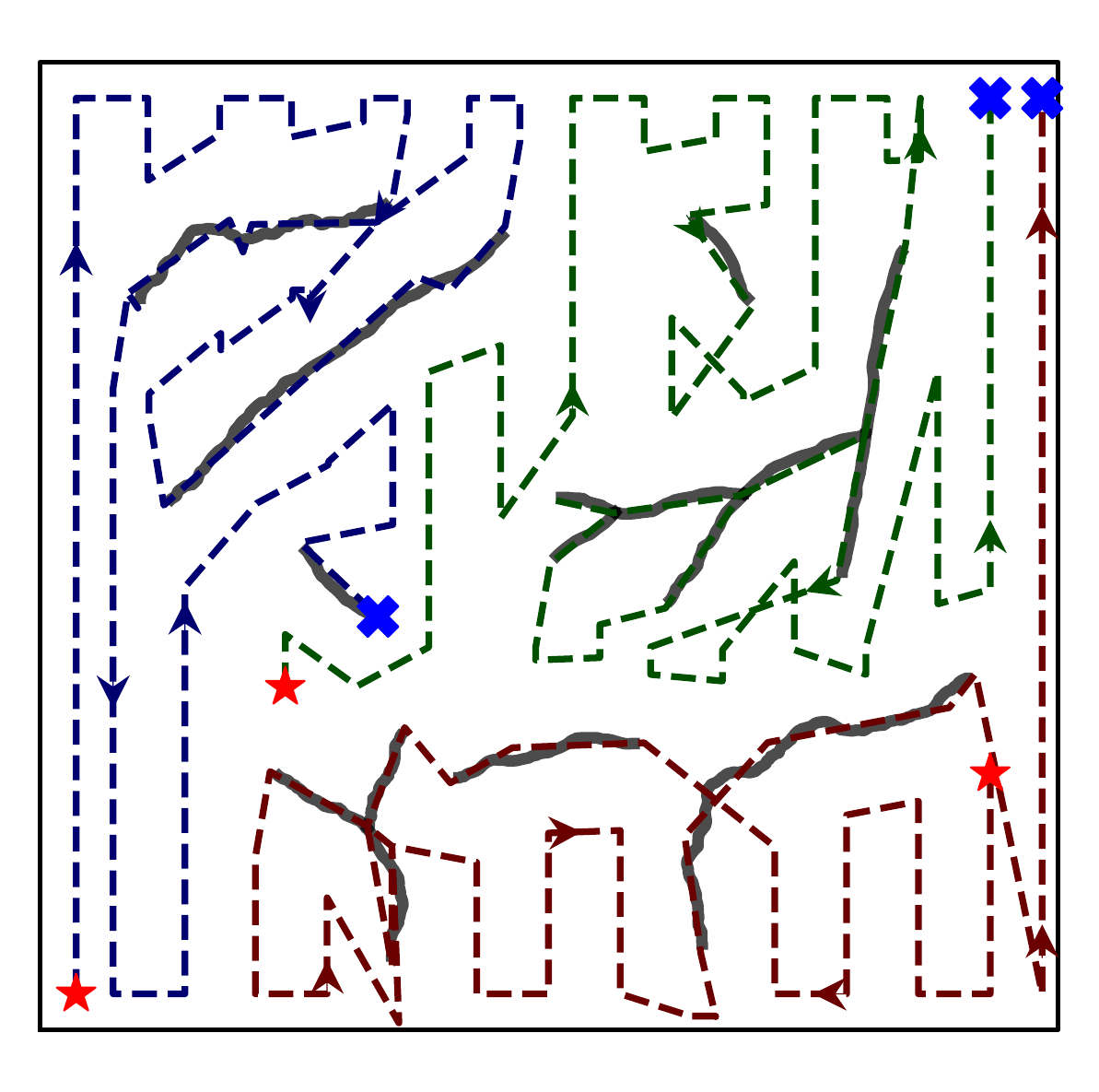}
    \end{subfigure}
    \begin{subfigure}[t]{0.24\textwidth}
        \centering
        \includegraphics[ width=\textwidth]{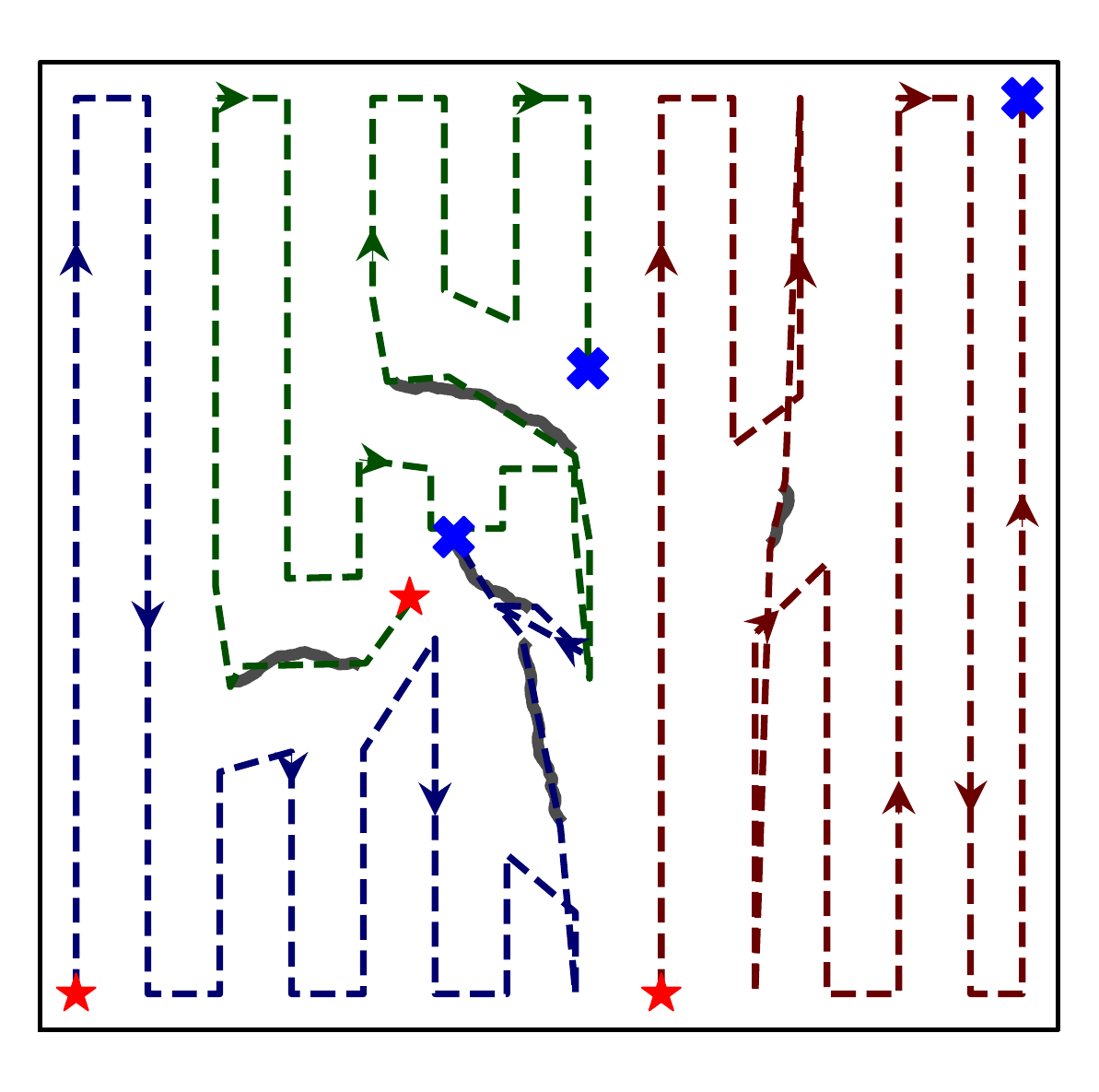}
    \end{subfigure}
    \begin{subfigure}[t]{0.24\textwidth}
        \centering
        \includegraphics[width=\textwidth]{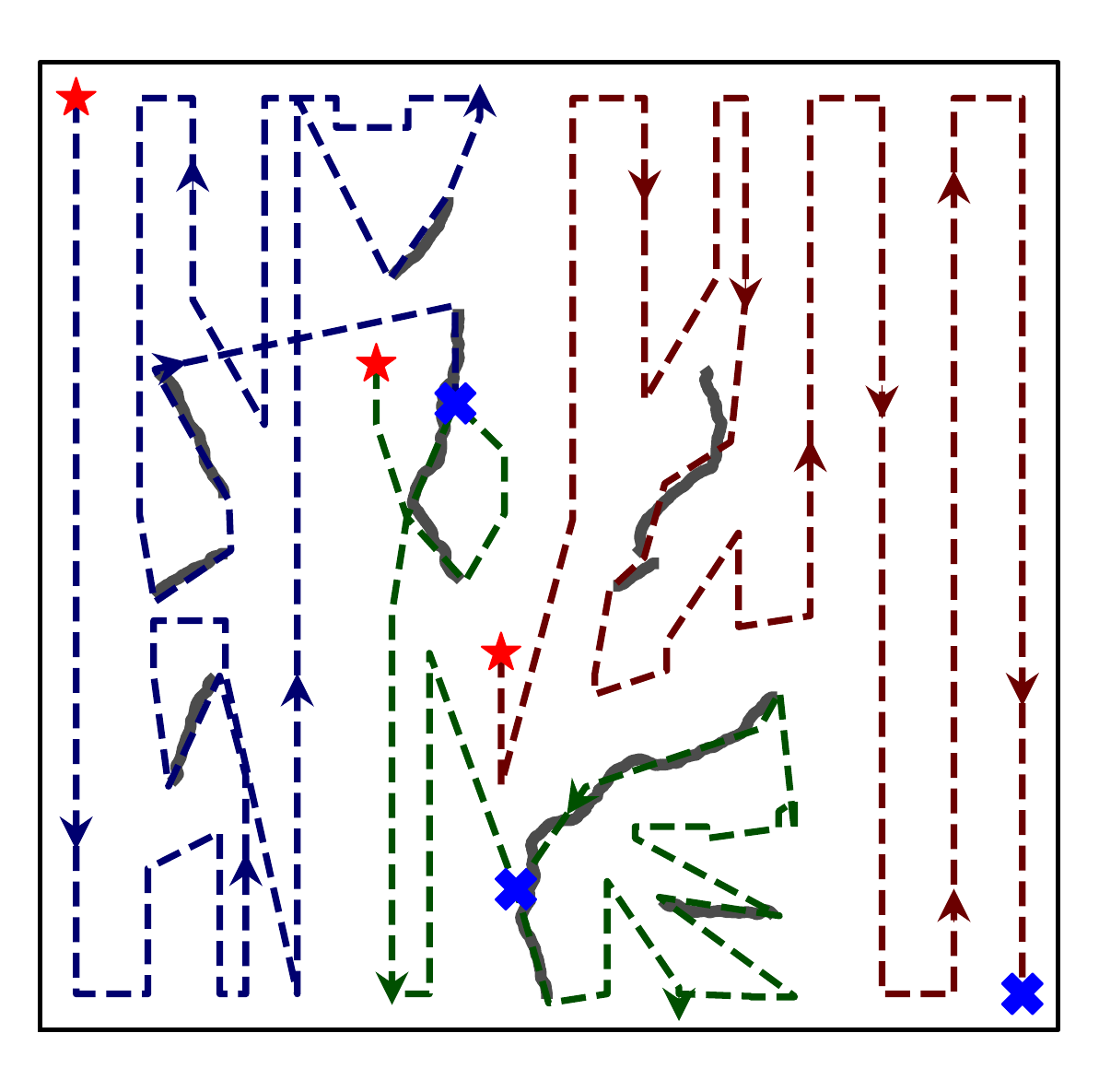}
    \end{subfigure}

    \begin{subfigure}[t]{0.24\textwidth}
        \centering
        \includegraphics[width=\textwidth]{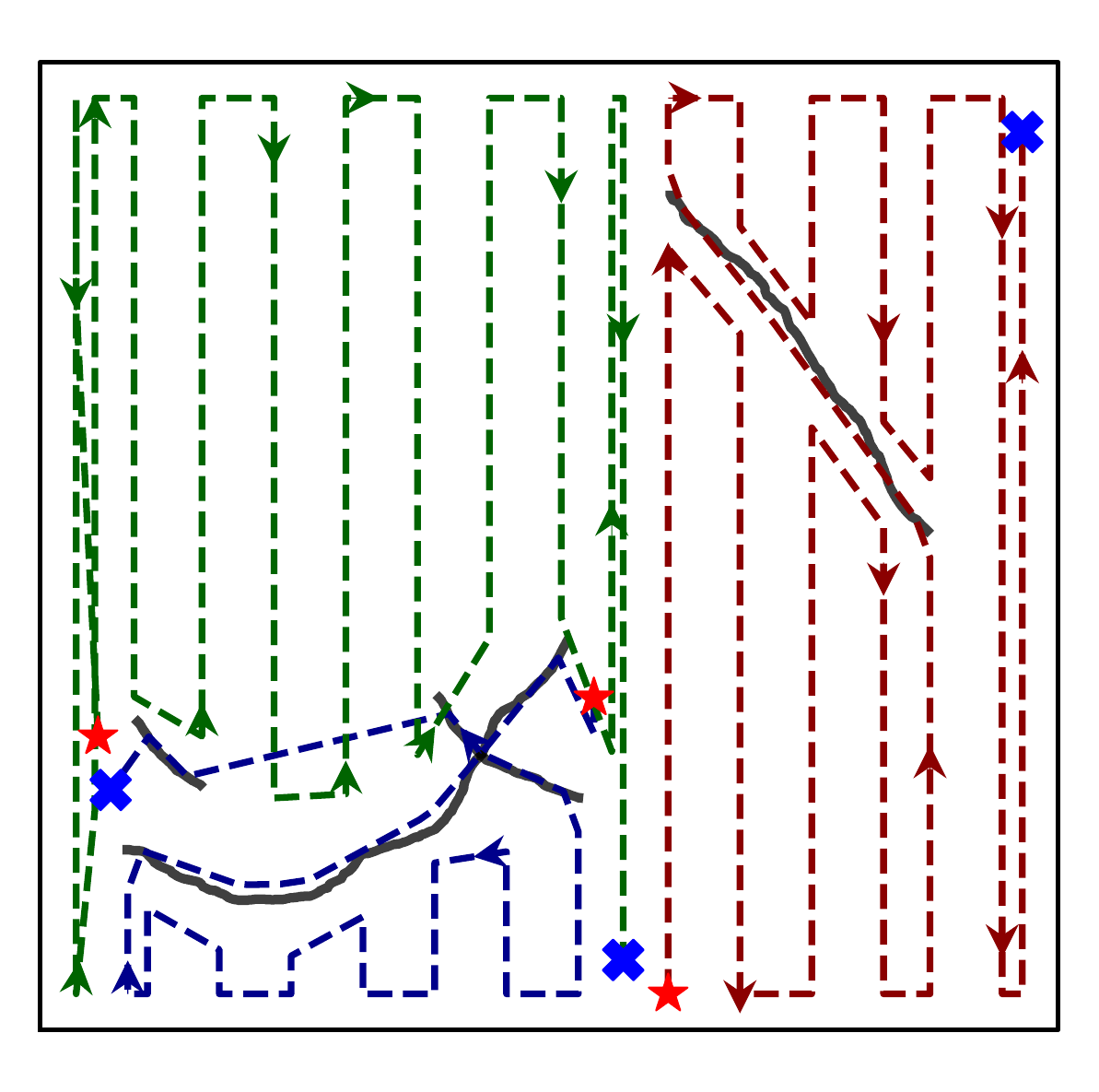}
        
    \end{subfigure}
    \begin{subfigure}[t]{0.24\textwidth}
        \centering
        \includegraphics[ width=\textwidth]{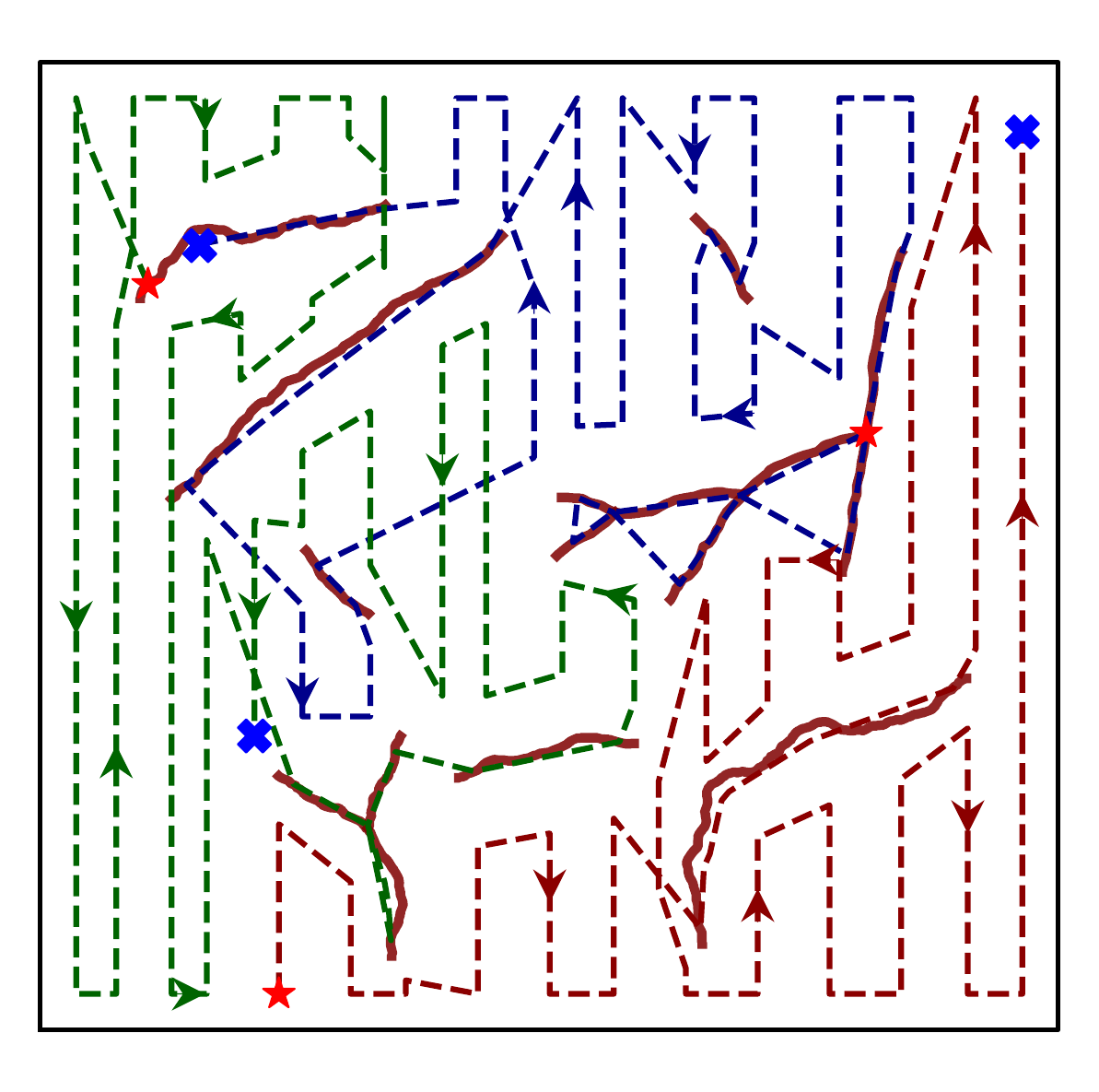}
        
    \end{subfigure}
    \begin{subfigure}[t]{0.24\textwidth}
        \centering
        \includegraphics[ width=\textwidth]{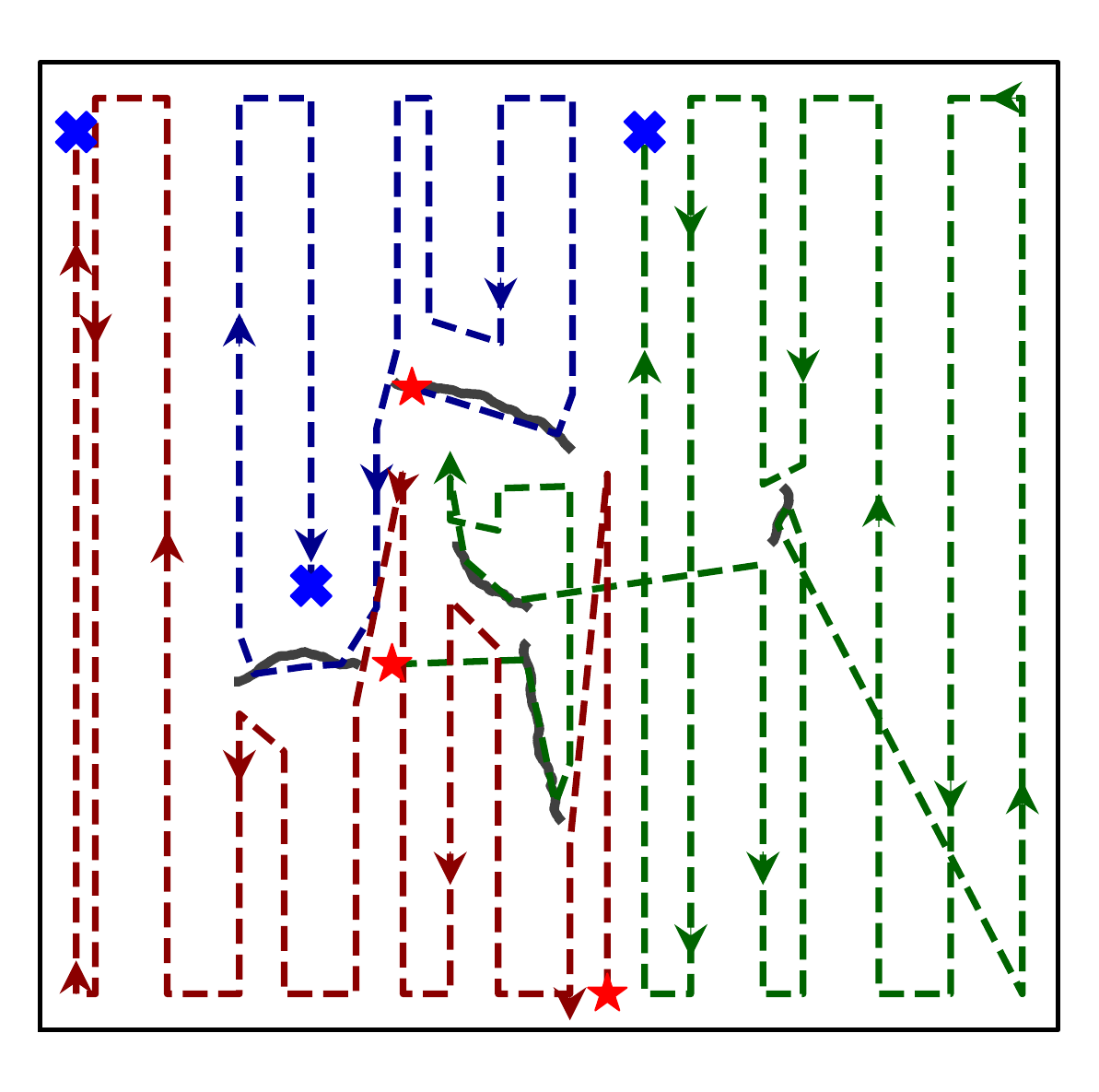}
        
    \end{subfigure}
    \begin{subfigure}[t]{0.24\textwidth}
        \centering
        \includegraphics[ width=\textwidth]{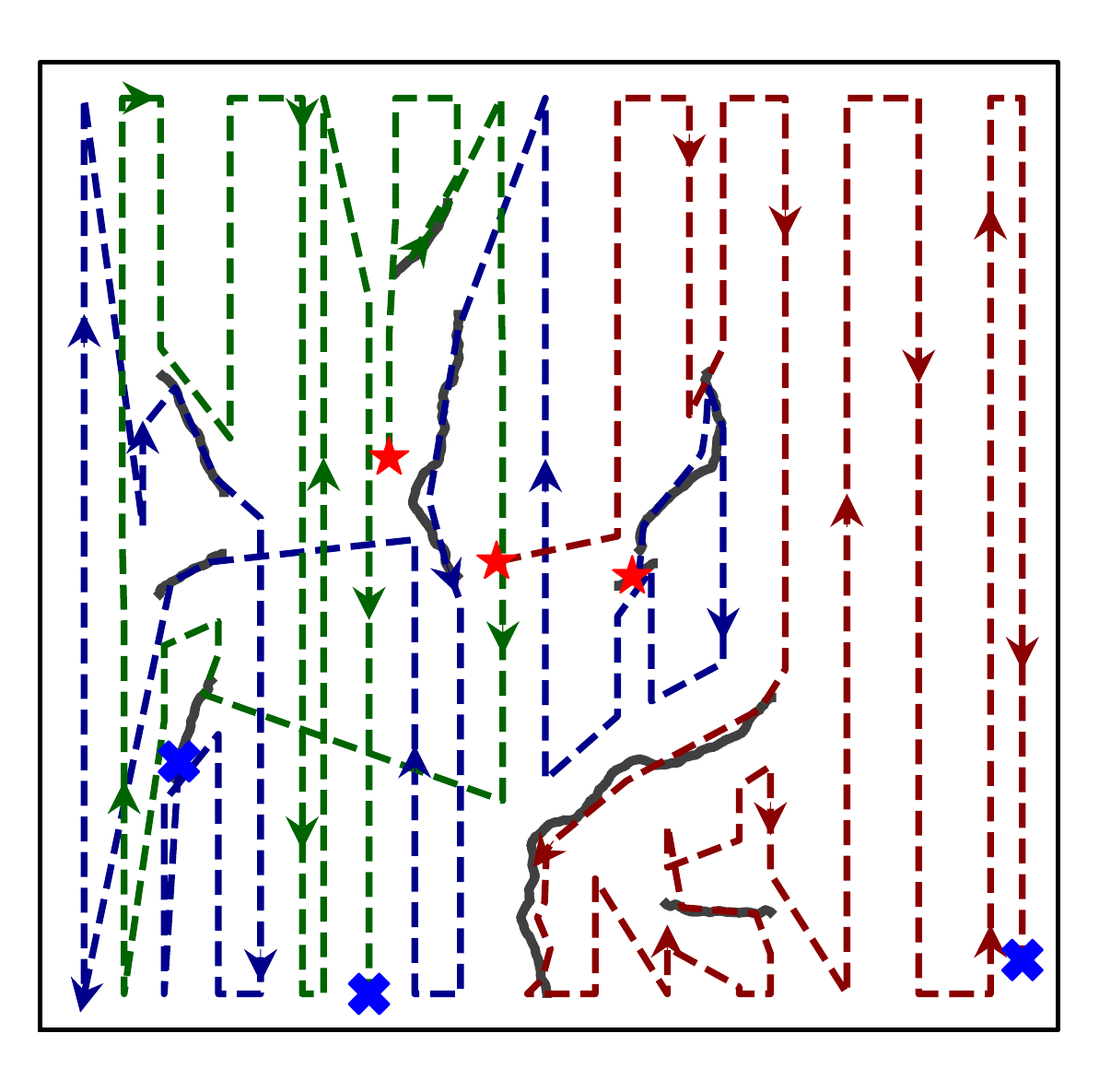}
        
    \end{subfigure}

    
    \begin{subfigure}[t]{0.24\textwidth}
        \centering
        \includegraphics[ width=\textwidth]{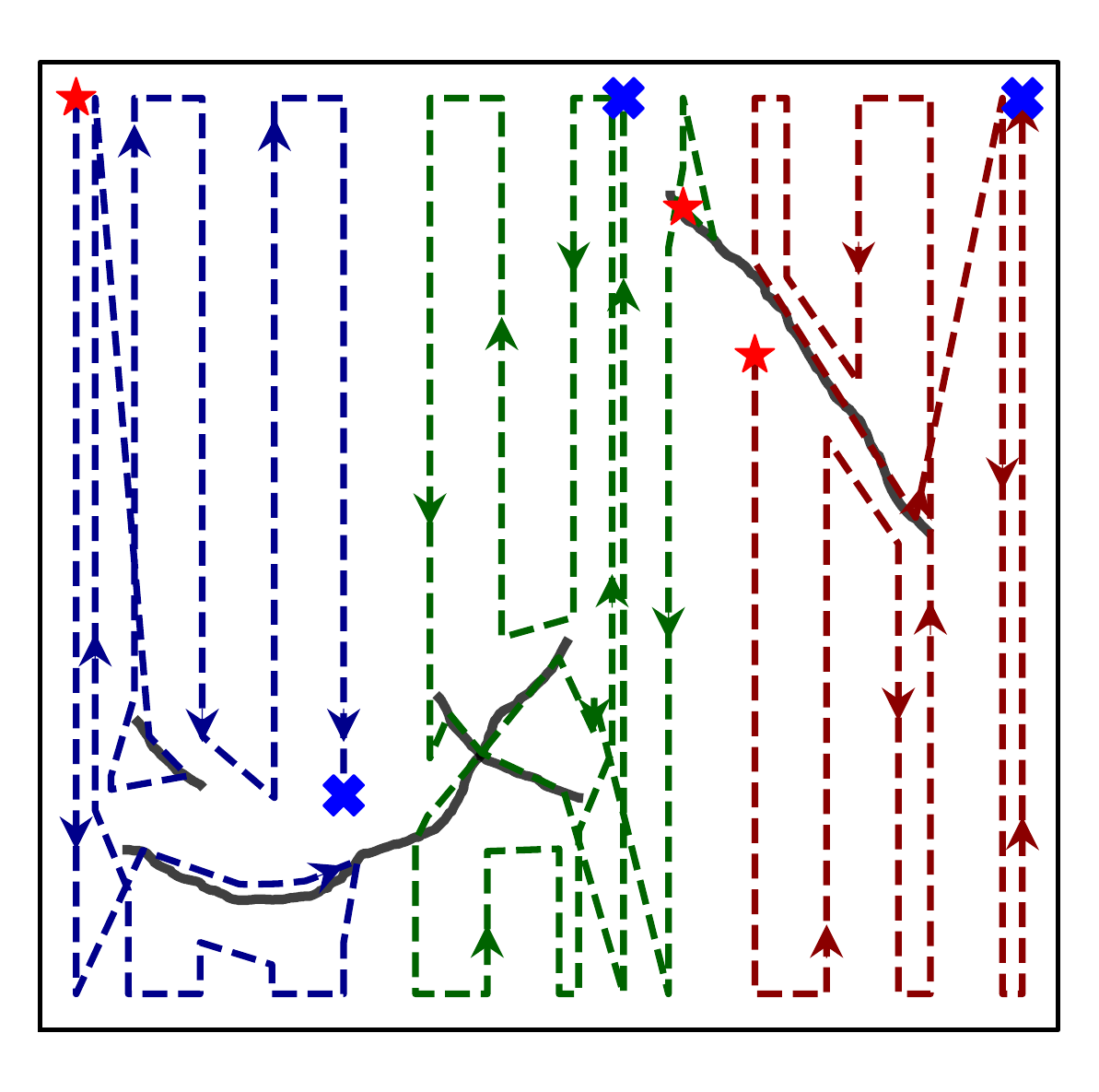}
        
    \end{subfigure}
    \begin{subfigure}[t]{0.24\textwidth}
        \centering
        \includegraphics[ width=\textwidth]{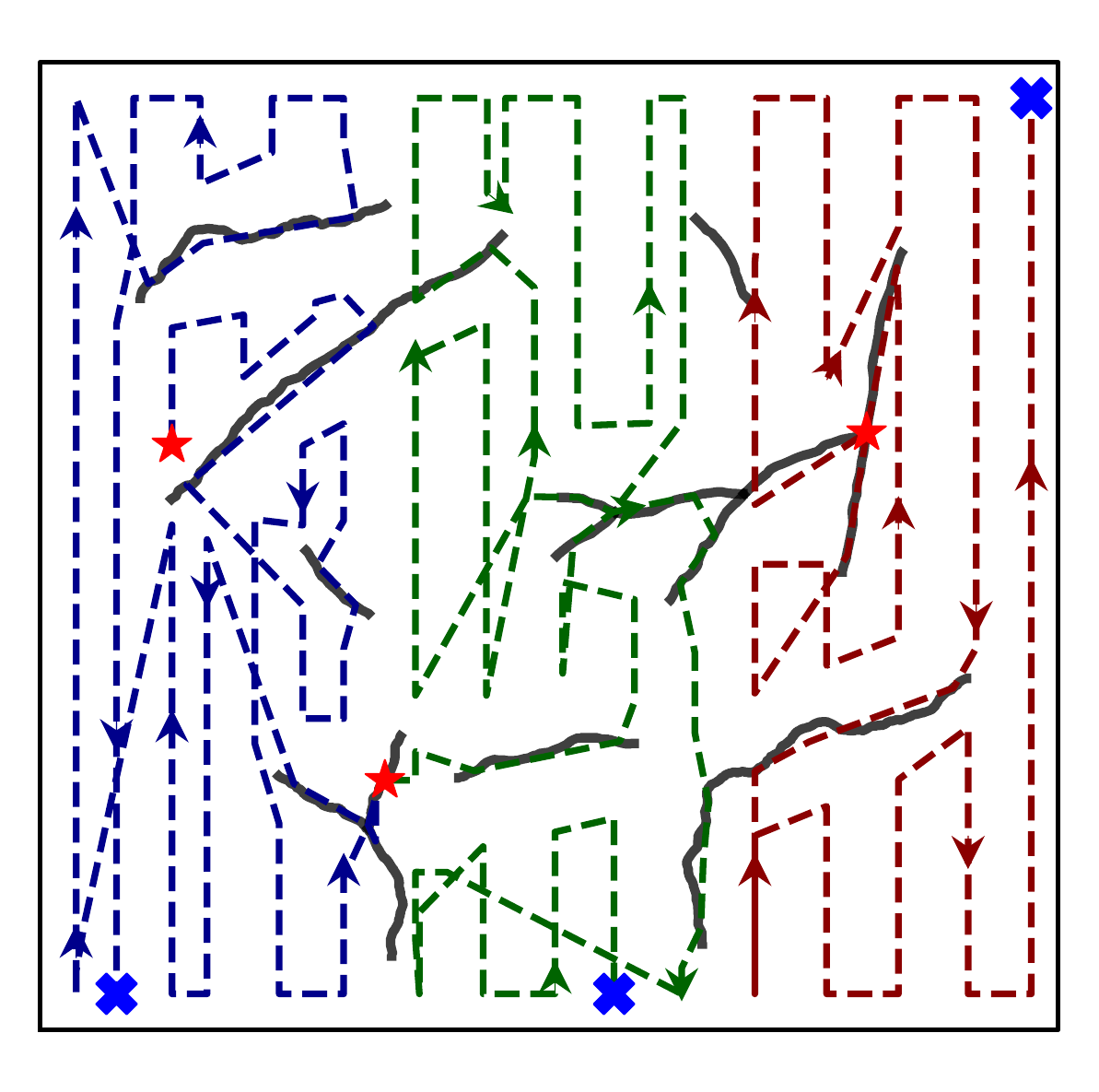}
        
    \end{subfigure}
    \begin{subfigure}[t]{0.24\textwidth}
        \centering
        \includegraphics[ width=\textwidth]{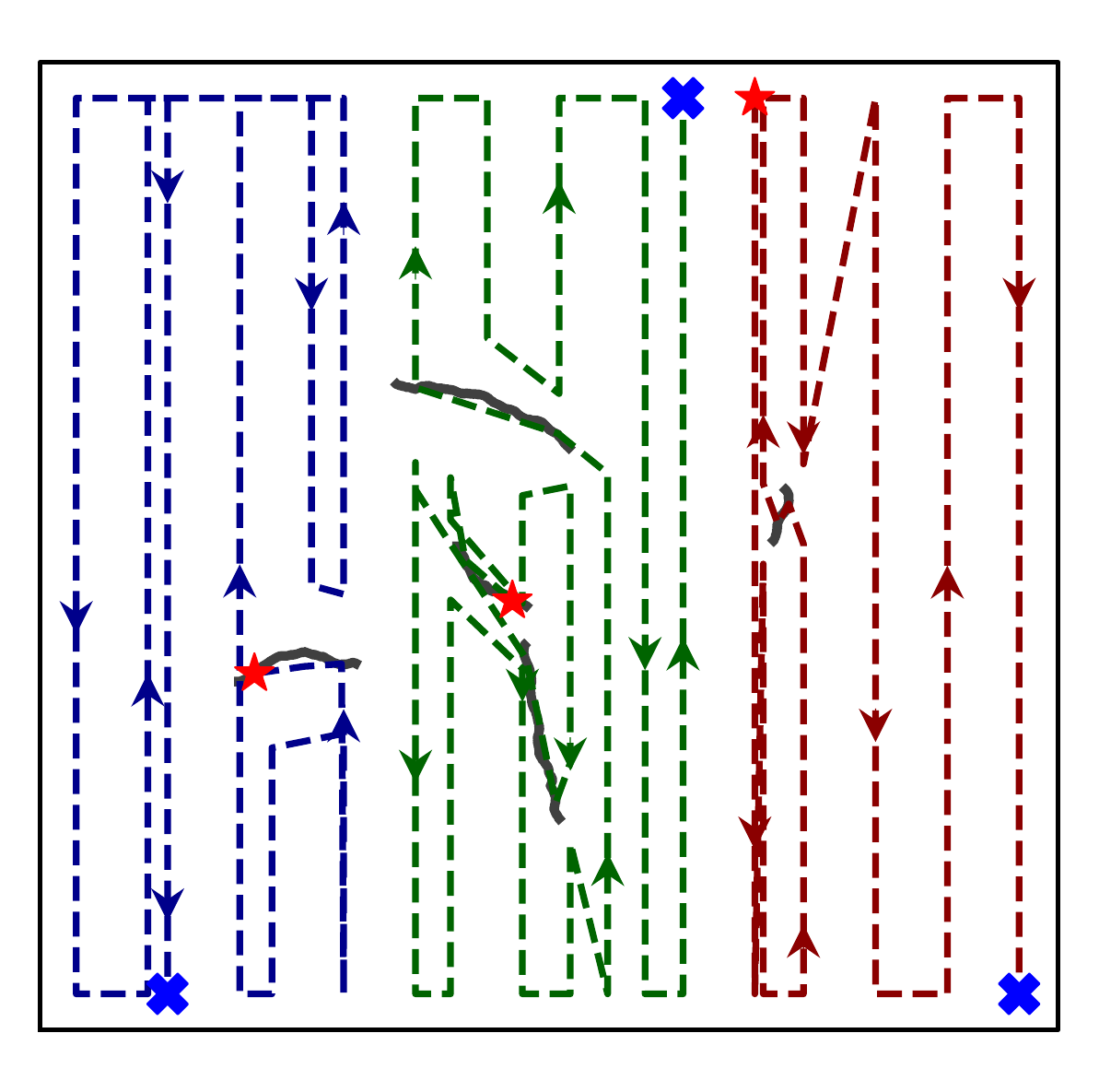}
        
    \end{subfigure}
    \begin{subfigure}[t]{0.24\textwidth}
        \centering
        \includegraphics[ width=\textwidth]{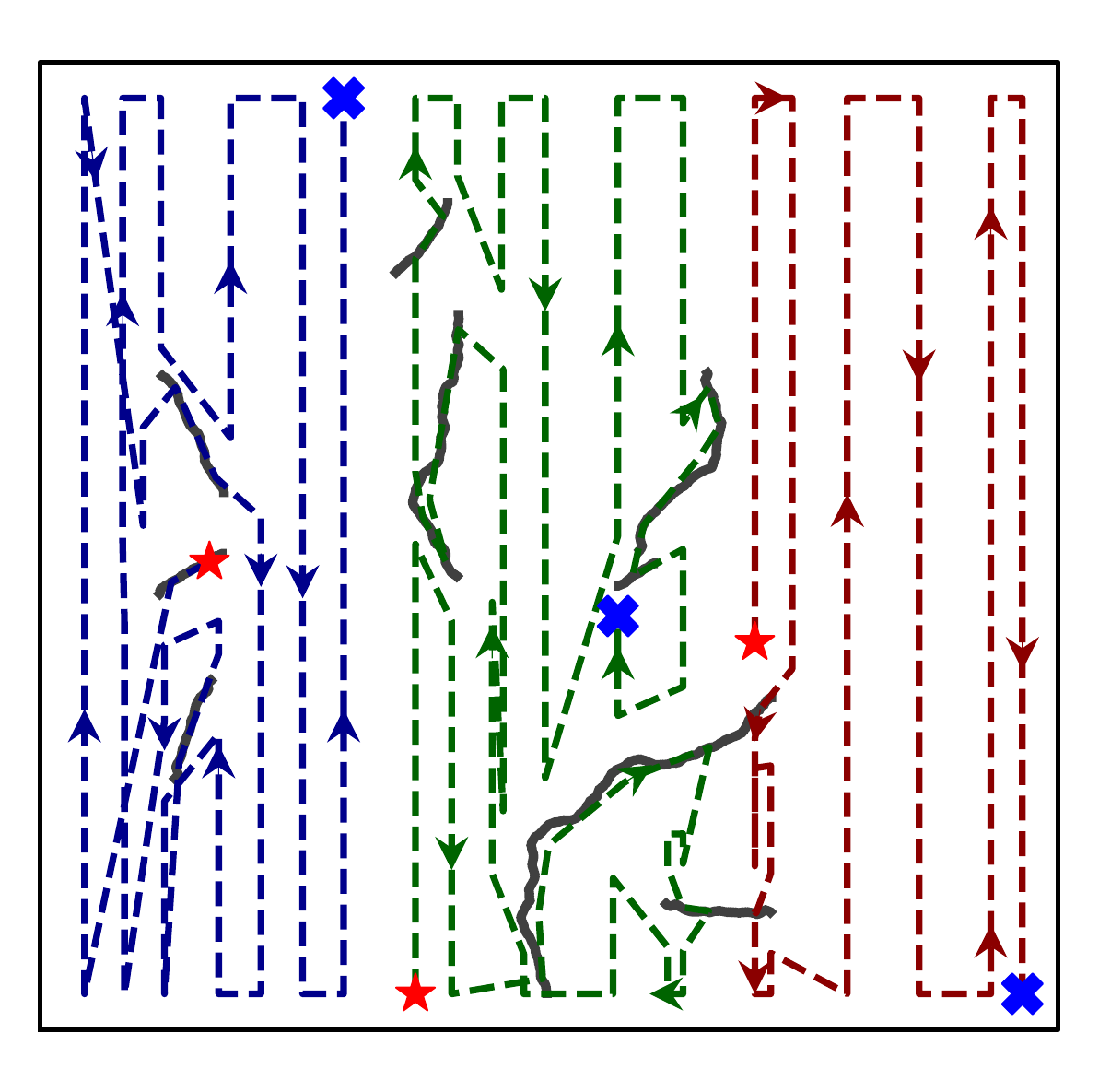}
        
    \end{subfigure}
    \begin{subfigure}[t]{0.24\textwidth}
        \centering
        \includegraphics[ width=\textwidth]{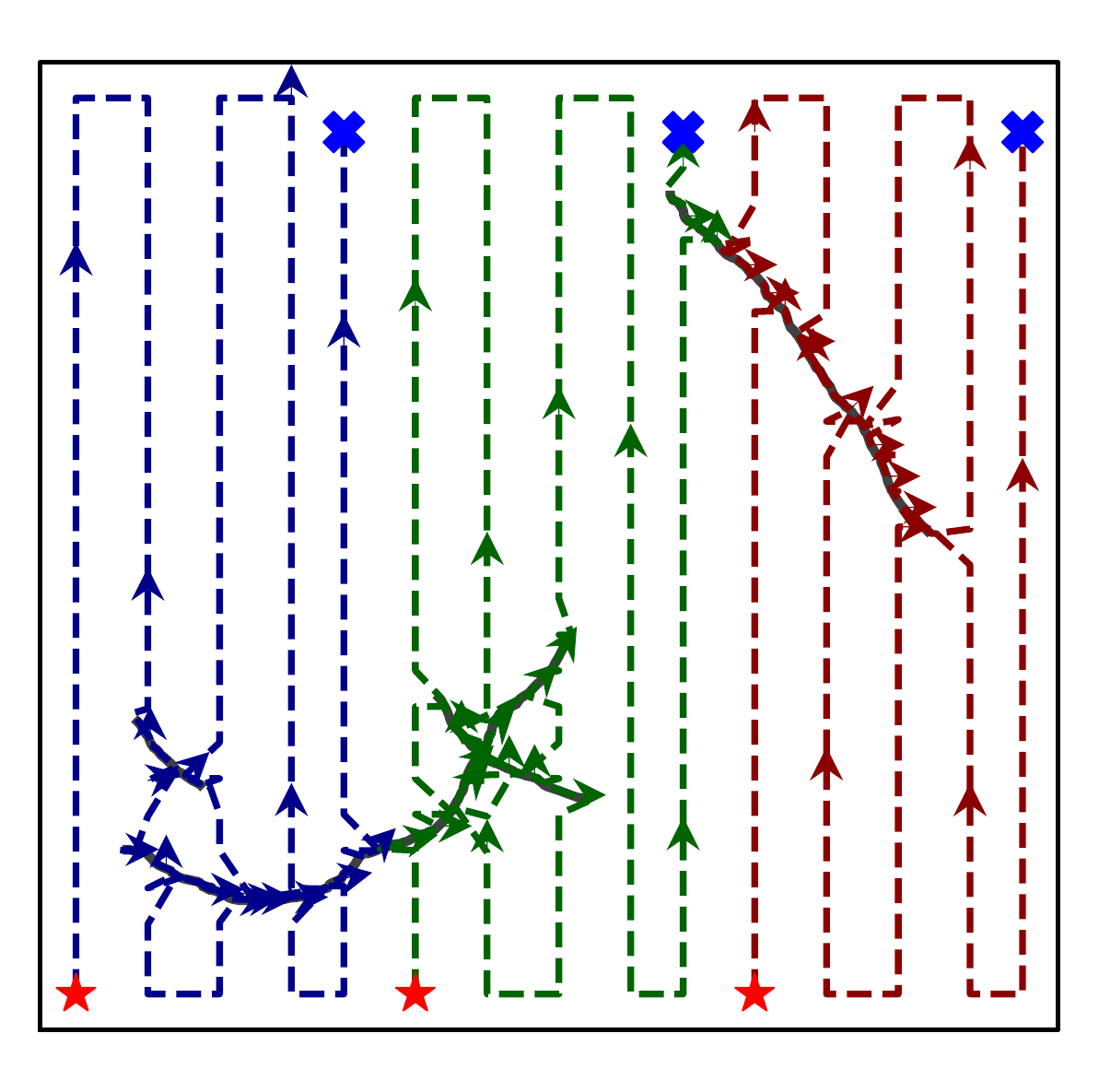}
        \caption{}
    \end{subfigure}
    \begin{subfigure}[t]{0.24\textwidth}
        \centering
        \includegraphics[ width=\textwidth]{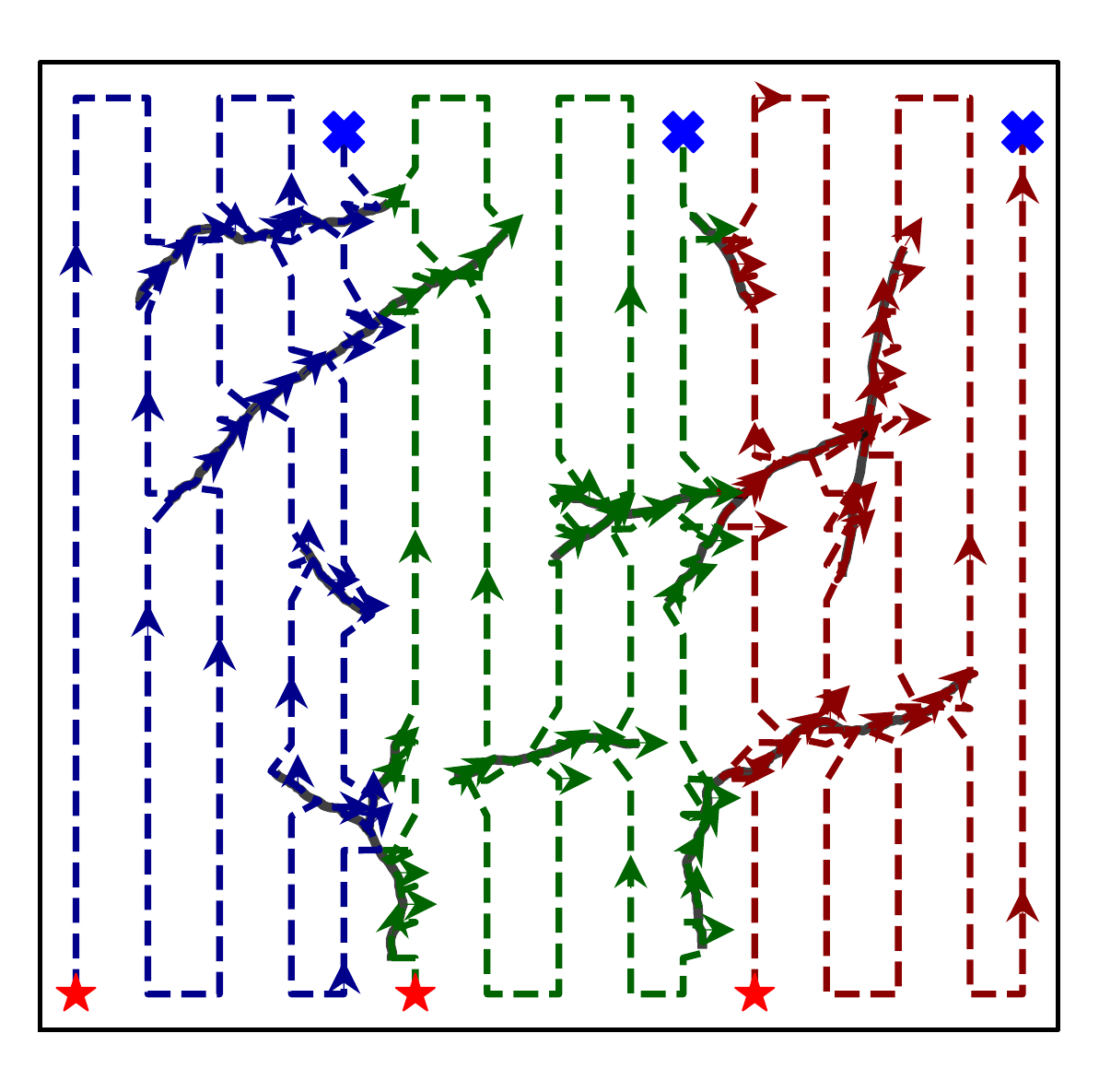}
        \caption{}
    \end{subfigure}
    \begin{subfigure}[t]{0.24\textwidth}
        \centering
        \includegraphics[ width=\textwidth]{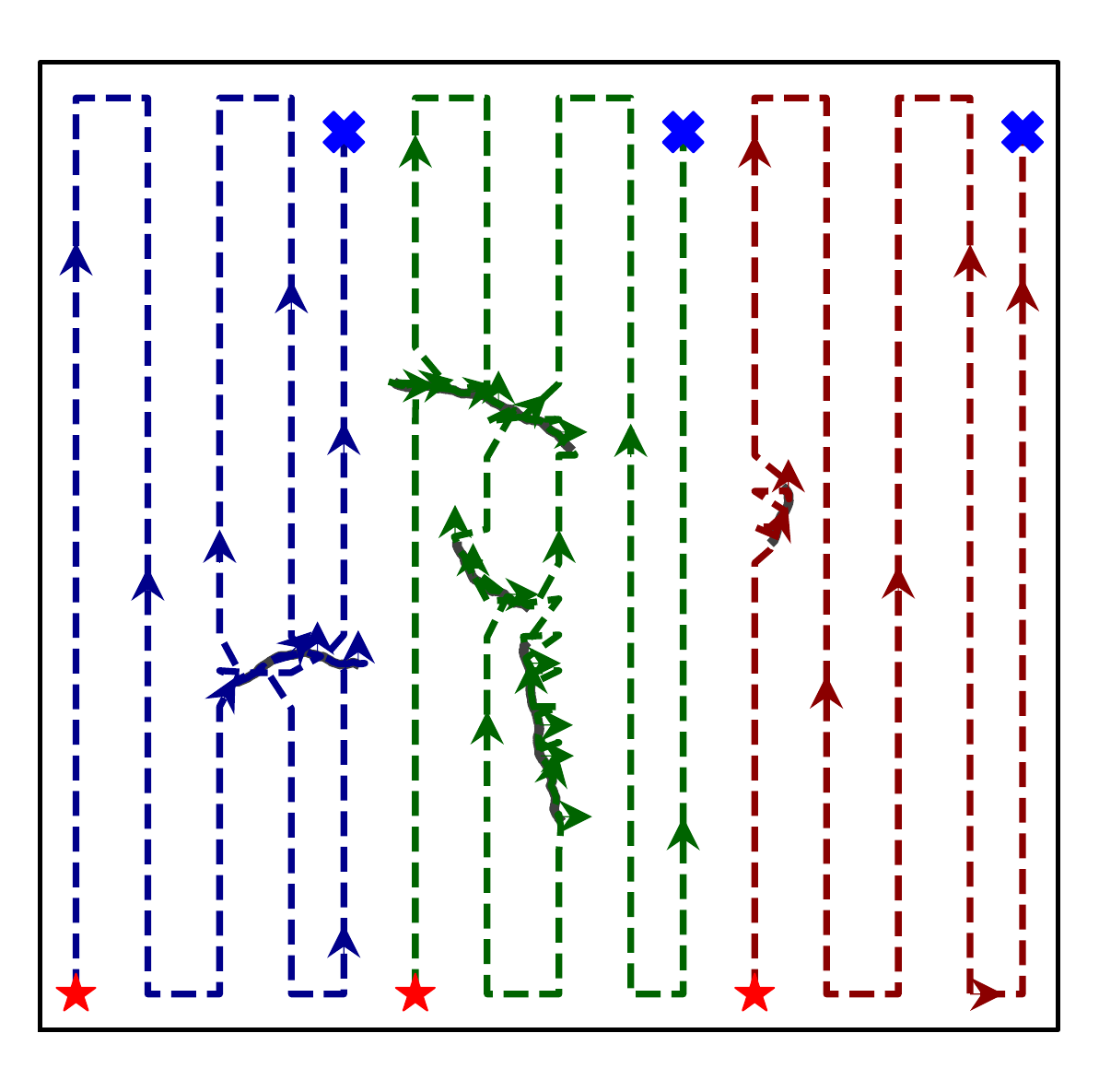}
        \caption{}
    \end{subfigure}
    \begin{subfigure}[t]{0.24\textwidth}
        \centering
        \includegraphics[ width=\textwidth]{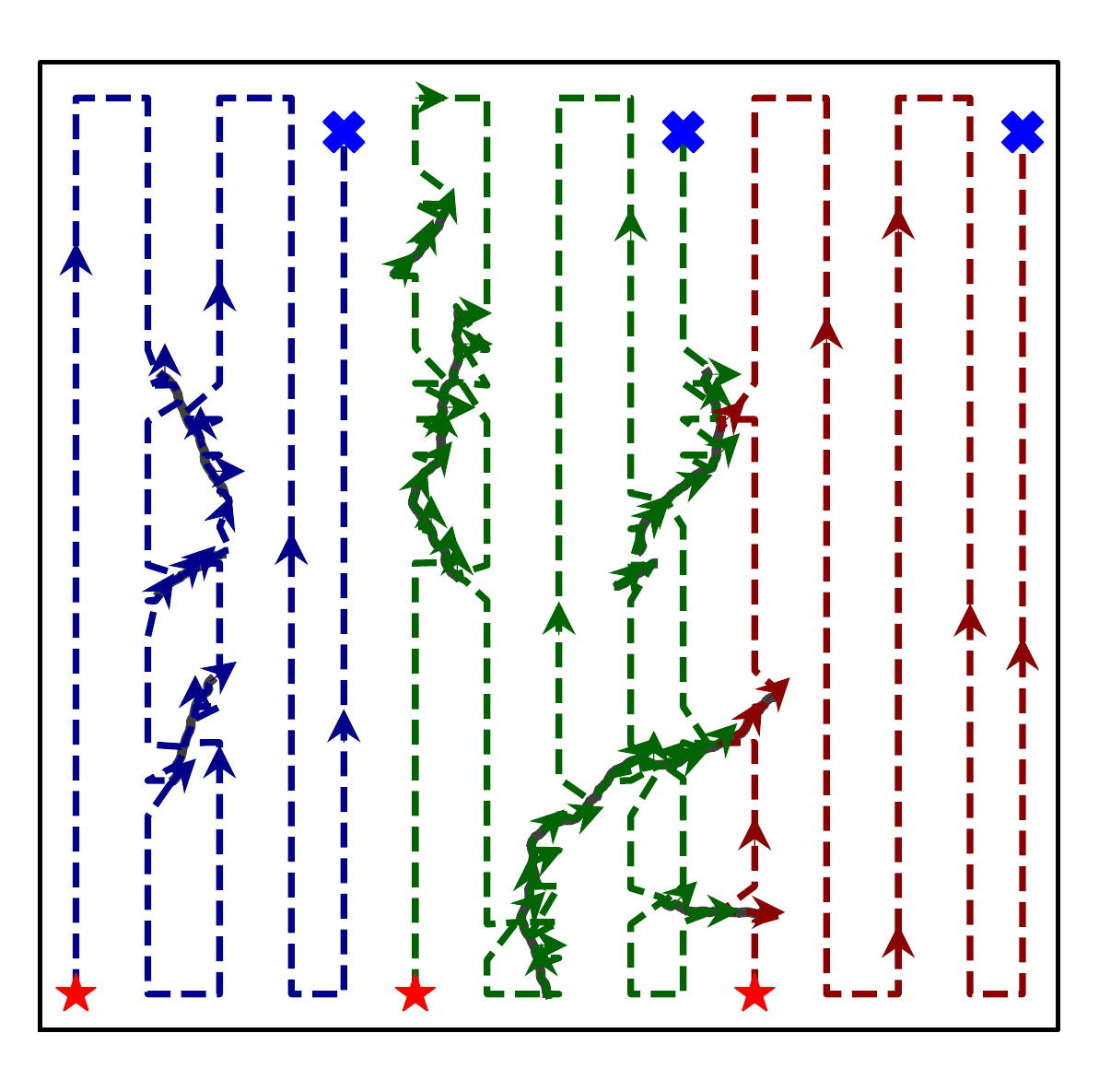}
        \caption{}
    \end{subfigure}

    \caption{Comparison of outcomes for three robots performing double coverage tasks across eight crack density and distribution maps. Each row represents the simulation results under one motion planning algorithm. From top to bottom, the results are under HCMR planner, G-mSCC planner, S-mSCC planner and mGreedy planner. From left to right, the columns represent crack maps in categories U65, U100, G65, G100, respectively. }
    \label{fig:three_robot_demo}
\end{figure*}

\begin{table*}[h!]
\centering
\caption{
Experiment Performance Comparison for Three-Robot Double Coverage On Eight Crack Maps}
\begin{tabularx}{\textwidth}{@{}l*{4}{>{\centering\arraybackslash}X}*{4}{>{\centering\arraybackslash}X}*{4}{>{\centering\arraybackslash}X}*{4}{>{\centering\arraybackslash}X}*{4}{>{\centering\arraybackslash}X}@{}}
\toprule
\textbf{ } 
& \multicolumn{4}{c}{\textbf{Total Path Length (\bs{m})}} 
& \multicolumn{4}{c}{\textbf{Robot Travel Time (s)} } 
& \multicolumn{4}{c}{\textbf{Sensor Coverage (\%)}} 
& \multicolumn{4}{c}{\textbf{Filling Coverage (\%)}} 
& \multicolumn{4}{c}{\textbf{Sensing Conflicts}}\\ 
\cmidrule(lr){2-5} \cmidrule(lr){6-9} \cmidrule(lr){10-13} \cmidrule(lr){14-17} \cmidrule(lr){18-21} 
 Crack dist. 
 & U65 & U100 & G65 & G100 
 & U65 & U100 & G65 & G100  
 & U65 & U100 & G65 & G100  
 & U65 & U100 & G65 & G100 
 & U65 & U100 & G65 & G100 
\\ 
\midrule 
HCMR 
& \textbf{407} & \textbf{424} & \textbf{462} & \textbf{466} 
& \textbf{145} & \textbf{145} & \textbf{161} & \textbf{148} 
& \textbf{99} & \textbf{103} & \textbf{113} & \textbf{113} 
& \textbf{100} & \textbf{100} & \textbf{100} & \textbf{100} 
& \textbf{0} & \textbf{0} & \textbf{0} & \textbf{0}
\\ 
G-mSCC 
& 486 & 554 & 523 & 545 
& 155 & 214 & 173 & 198 
& 118 & 135 & 127 & 133 
& 94 & 96 & 98 & 97
& 0 & 5 & 6 & 5 
\\ 

S-mSCC
& {567} & {548} & {567} & {594}
& {183} & {197} & {187} & {197} 
& {138} & {133} & {138} & {145} 
& {96} & {85} & 98 & {91} 
& \textbf{0} & \textbf{0} & \textbf{0} & \textbf{0} 
\\ 

mGreedy
& 502 & 584 & 468 & 512 
& 180 & 231 & 174 & 238 
& 122 & 142 & 114 & 124 
& \textbf{100} & \textbf{100} & \textbf{100} & \textbf{100} 
& \textbf{0} & 1 & \textbf{0} & 1
\\ 

\bottomrule
\label{table:three_robot_demo}
\end{tabularx}
\end{table*}

\subsection{Multi-Agent Allocation and Trajectory Generation}
For each Eulerian tour $p \in \mathcal{P}$, we apply a balanced partitioning algorithm in~\cite{Xu-2011-7344} to distribute the workspace among multiple robots. This algorithm has a time complexity of $O(|E| \log W)$ and a space complexity of $O(|E|)$, where $W$ represents the total edge weight. This approach ensures an efficient and balanced allocation of tasks among the robots. It is worth noting that the complexity is independent of the number of robots. For all walks in $\mathcal{P}$, the total time complexity is $O(|\mathcal{P}| |E| \log W)$. Then we choose the partition that achieves (\ref{eq.objective}).  In this way, each robot is assigned with a balanced subpath of the Eulerian tour. 

For each subpath, we leverage the Morse boundedness structure of our construction. 
Adjacent edges in $\bs{E}_w$ are Morse invariant at their stitching points, enabling 
efficient trajectory generation as described in Algorithm~\ref{algo:generate_total_trajectory}. 
Specifically, starting from an edge in $\bs{E}_w$, we iteratively and greedily merge 
the corresponding cells ($\phi^{-1}$) until reaching an edge whose successor no longer belongs to the Reeb graph (lines 4--7). The function \textsc{Boustrophedon} then optimizes the Boustrophedon path for the merged polygon (\textit{mergedPoly}) exactly once (line 8). Moreover, the optimality of the current Boustrophedon path does not influence subsequent trajectory generations. This approach eliminates the need to optimize local zig-zag patterns within an exponentially growing search space, as seen in \cite{xie2018integrated}, or to rely on heuristics, as in \cite{tung2019solution}.

\begin{algorithm}[t]
\DontPrintSemicolon
\caption{Generate Trajectory}\label{algo:generate_total_trajectory}
\SetAlgoVlined
\SetKwInOut{Input}{Input}
\SetKwInOut{Output}{Output}
\Input{$\mathcal{E}$, $\mathbb{G}$}
\Output{$T$}

$T \gets [\,]$, $\text{reebEdges} \gets [\,]$, $\text{mergedPoly} \gets \emptyset$\;

\For{$i \gets 1$ \KwTo $|\mathcal{E}|$}{
    $label \gets$ \Call{GetEdgeLabel}{$\mathcal{E}[i], \mathbb{G}$}\;
    \uIf{$label = \text{`reeb'}$}{
        $\text{reebEdges} \gets \text{reebEdges} + [\mathcal{E}[i]]$\;
        $\text{mergedPoly} \gets \text{mergedPoly} \cup \phi^{-1}(\mathcal{E}[i])$\;

        \If{next edge $\notin \mathcal{E}_w$ \textbf{or} $i = |\mathcal{E}|$}{
            $T \gets$ \Call{Boustrophedon}{$\text{mergedPoly}$}\;
            $\text{reebEdges} \gets [\,]$, $\text{mergedPoly} \gets \emptyset$\;
        }
    }
    \ElseIf{$label = \text{`crack'}$}{
        $T \gets$ \Call{CrackFollow}{$\mathcal{E}[i]$}\;
    }
}
\end{algorithm}

We further have the collision-free property by this construction:
\begin{thm}\normalfont \label{lem:collision_free}
  The trajectories generated for different robots under the framework 
  satisfy the collision-free constraints 
  \eqref{eq.area_collision_constraint} 
  and \eqref{eq.crack_collision_constraint}, 
  except possibly at odd node pairings in $\mathbb{G}_{\text{aug}}$.
\end{thm}

\begin{pf}
Odd node pairings do not contribute to any manifold. Morse boundedness ensures that each robot’s coverage manifold 
is {simply connected relative to} the others, so they cannot 
overlap in their interior regions. 
\end{pf}

\begin{figure*}[h!]
    \centering
        \includegraphics[ trim=0cm 1.7cm 0cm 0cm, clip,width=\textwidth]{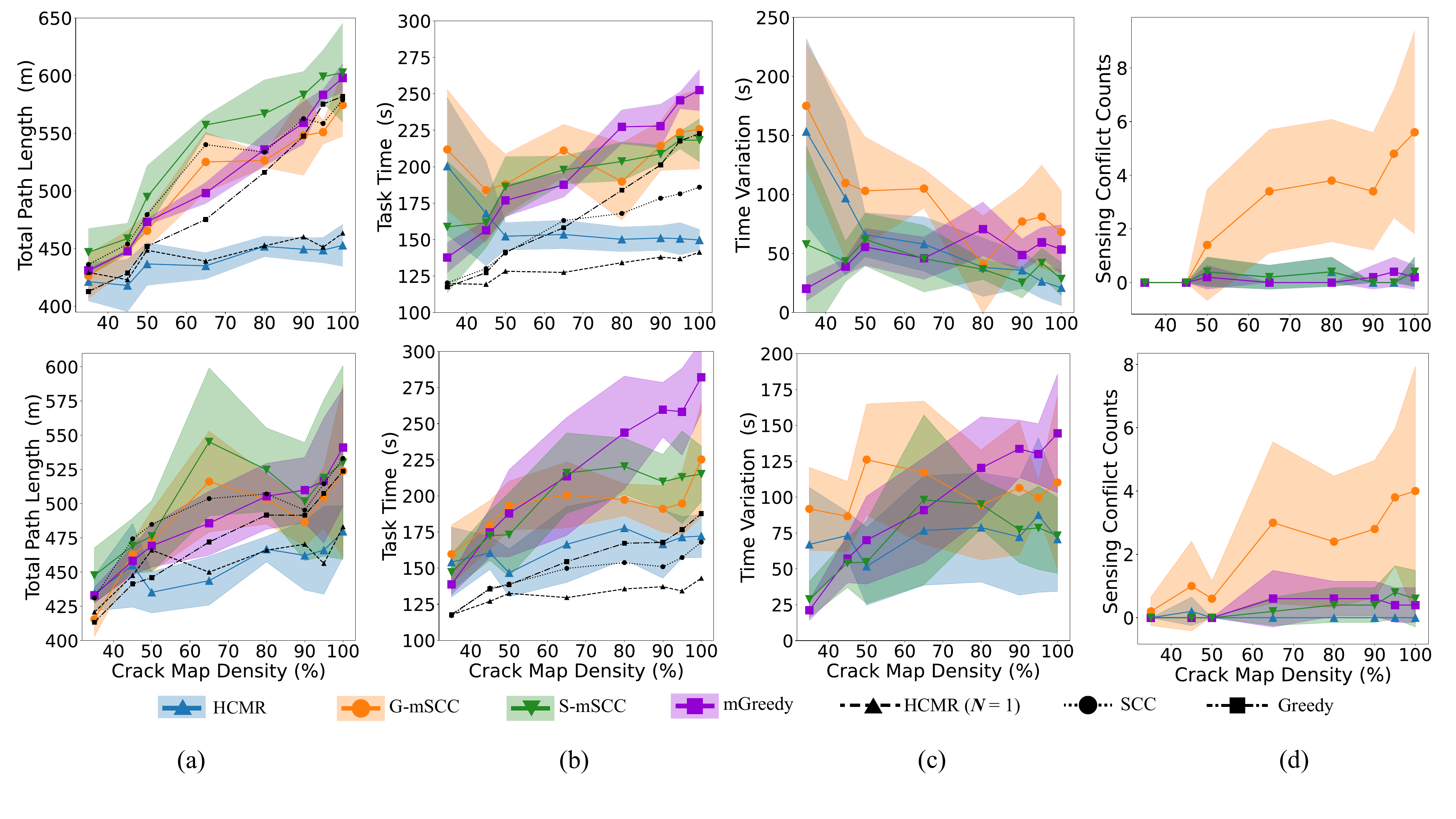}       
\caption{Comparison of evaluation metrics for different planning algorithms on uniformly distributed and Gaussian-distributed cracks with varying densities in the free space. The upper row presents results for uniformly distributed cracks, while the lower row displays results for Gaussian-distributed cracks. (a) Total path length for double coverage, where black marks indicate the average path length for a single robot, calculated in simulation for each crack category. (b) Task completion time for double coverage, with black marks representing the average time for a single robot, calculated in simulation for each crack category and divided by the number of robots. (c) Time variation among robots during task assignments. (d) Sensing conflict counts among robot trajectories. The lines represent the mean values, while the shaded areas represent one
standard deviation.}
\label{fig:performance_statistics}
\end{figure*}

\begin{figure}[h!]
    \centering
    \begin{subfigure}[t]{0.241\textwidth}
        \centering
        \includegraphics[ trim=0cm 0.2cm 0cm 0cm, clip,width=\textwidth]{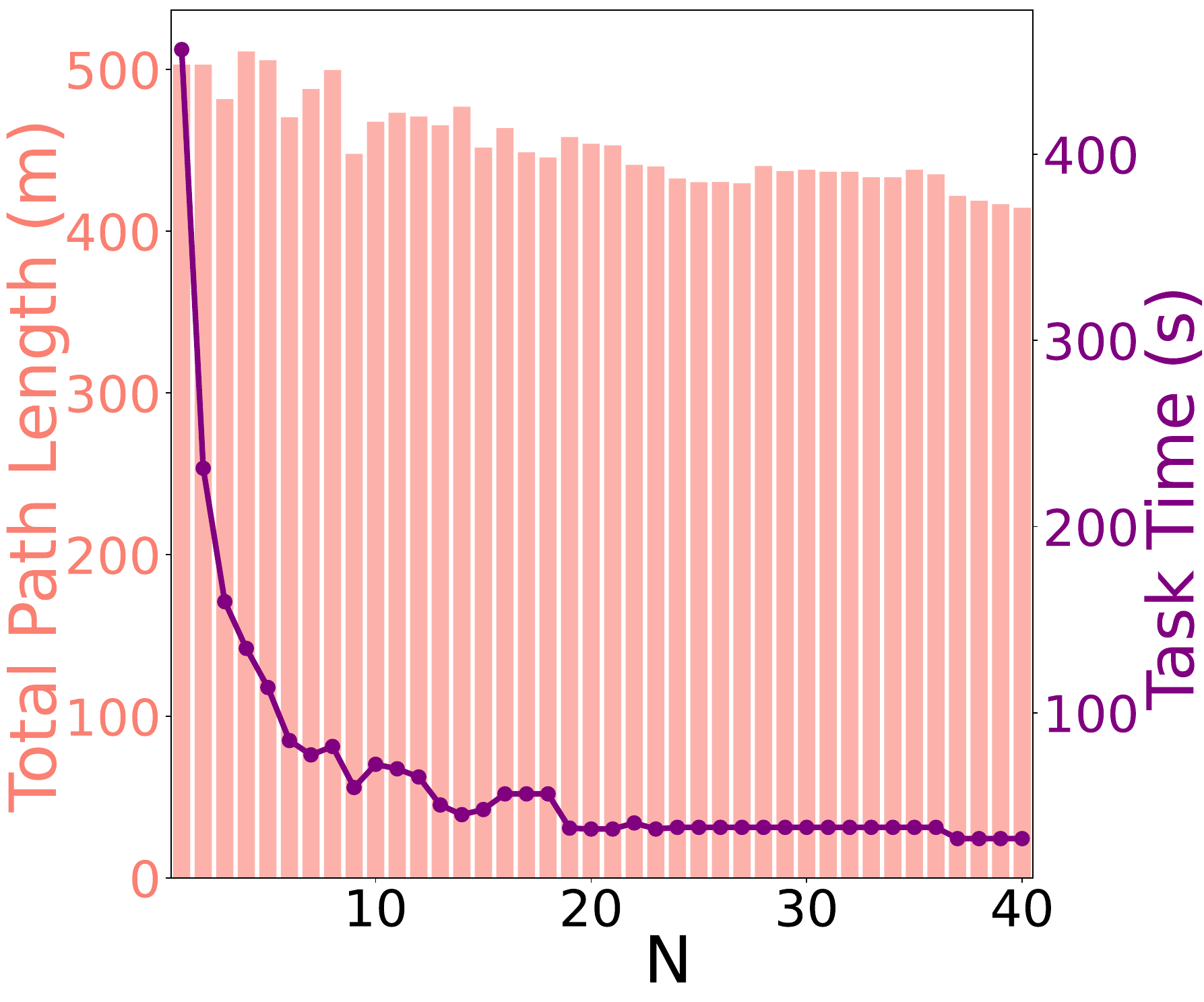}
        \caption{}
    \end{subfigure}
    \hfill
    \begin{subfigure}[t]{0.241\textwidth}
        \centering
        \includegraphics[trim=0cm 0.2cm 0cm 0cm, clip, width=\textwidth]{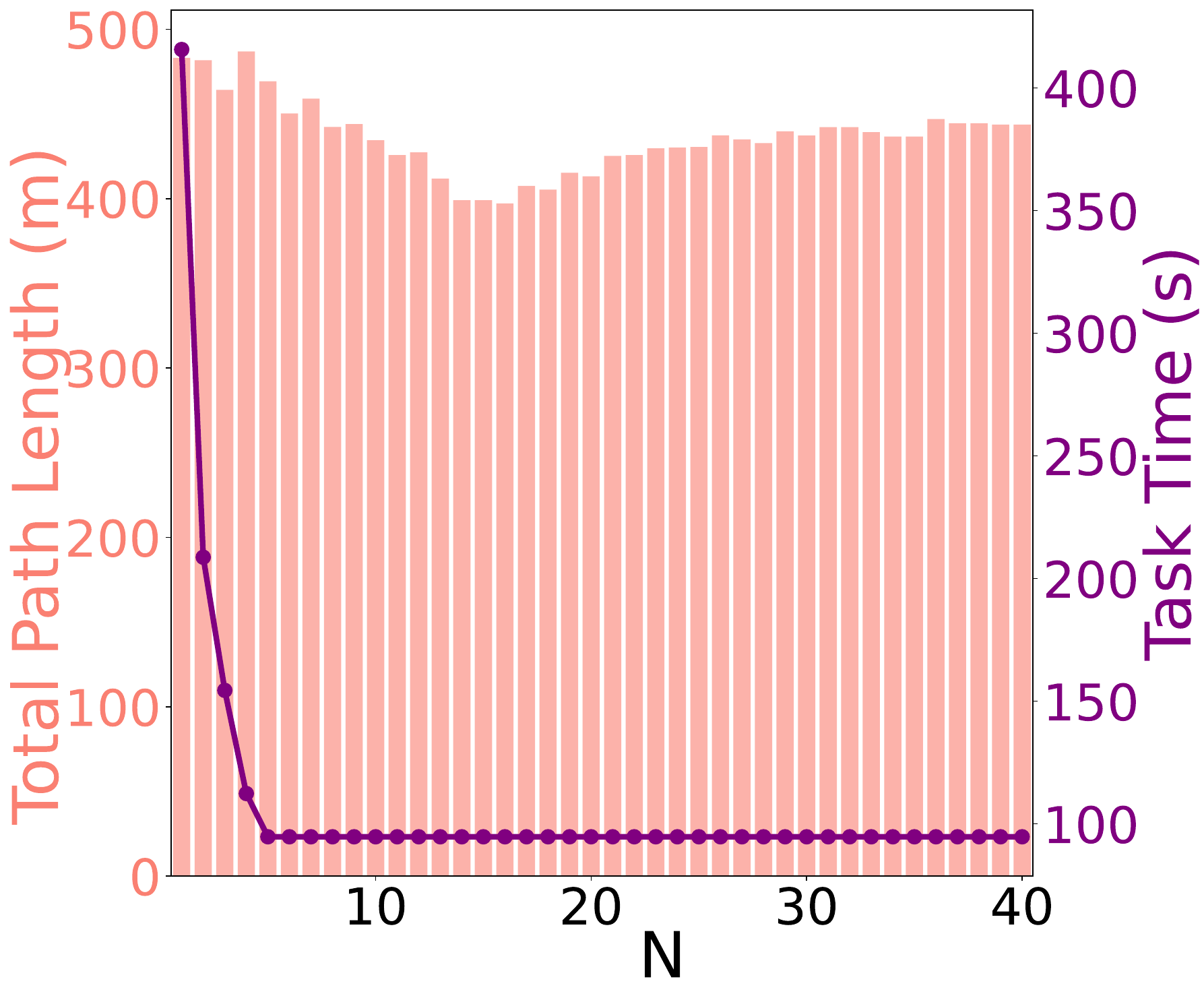}
        \caption{}
    \end{subfigure}
\caption{MDC path lengths and time costs for robot numbers ranging from 1 to 40 using the HCMR algorithm on the same map. (a) U100 (b) G100.}  

\label{fig:robot_num_statistics}
\end{figure}

\section{Simulation Results}
We conduct simulation studies to demonstrate the effectiveness of the proposed planning algorithm.
\subsection{Simulation Setup}
The algorithms for the MDC problem were implemented in \textsc{python}. We used \textsc{NetworkX} for graph management and basic grpah-related operations. The program is executed on a high-performance portable micro-processor (Intel NUC11PAHi7, Intel Corp.).

In the simulation, we considered a crack filling application with $l=30.5$~m and $w=29.0$~m. The robots need to fill the cracks on the ground as well as completely sense the region. Robots are equipped with omni wheels so turning costs are ignored. Each robot has a perception radius of $r=1.52$ m, and the filling nozzle operates within a radius of $a=0.44$~m. The speed each robot moves while sensing is set to $v_e=1.2$~m/s, and the speed while filling cracks is $v_s=0.4$ m/s.

To generate crack maps with varying densities and distributions, we combined cracks from the crack image database in \cite{shi2016automatic} with locations and orientations sampled from uniform and Gaussian distributions. The density of a crack (a branch $\bs{b}$) is metricized by the Minkowski sum area of $\bs{b} \oplus r$ over the workspace area. In this way, we categorize maps varies from $35\%$ to $100\% $.  with different distributions. 
Each category is denoted as [\text{distribution} + \text{density}], for example, U80 represents  uniformly distributed cracks with an 80\% density. For each category, we generated five maps for the simulations.

\begin{figure*}[h!]
    \centering

        \includegraphics[ trim=0cm 3cm 0cm 0cm, clip,width=\textwidth]{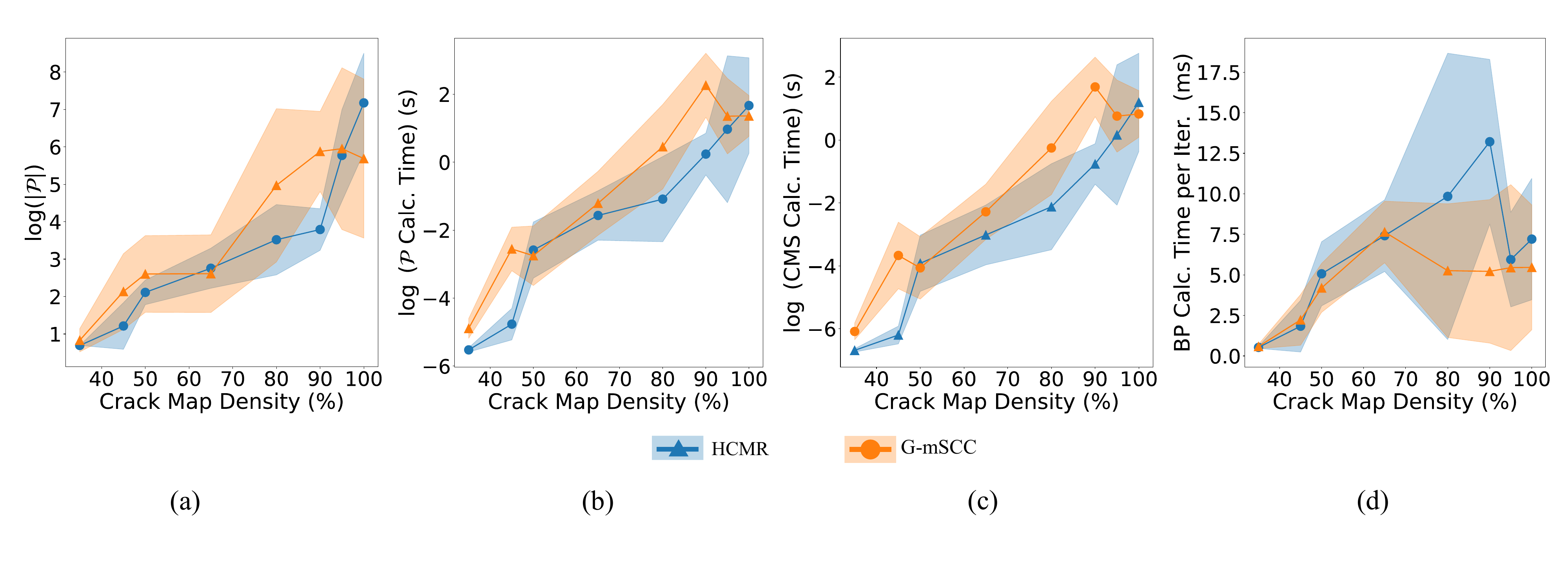}

\caption{Computational analysis for the HCMR algorithm. (a) The logarithm of the size of the Morse bounded collection, $|\mathcal{P}|$. (b) The logarithm of the computation time for determining $\mathcal{P}$. (c) The logarithm of the CMS process computation time. (d) The computation time for the BP process in each iteration. The lines represent the mean values, while the shaded areas represent one
standard deviation.}

\label{fig:computation_statistics}
\end{figure*}
\begin{figure}[t]
    \centering

        \includegraphics[ trim=0cm 1.8cm 0cm 0cm, clip,width=0.5\textwidth]{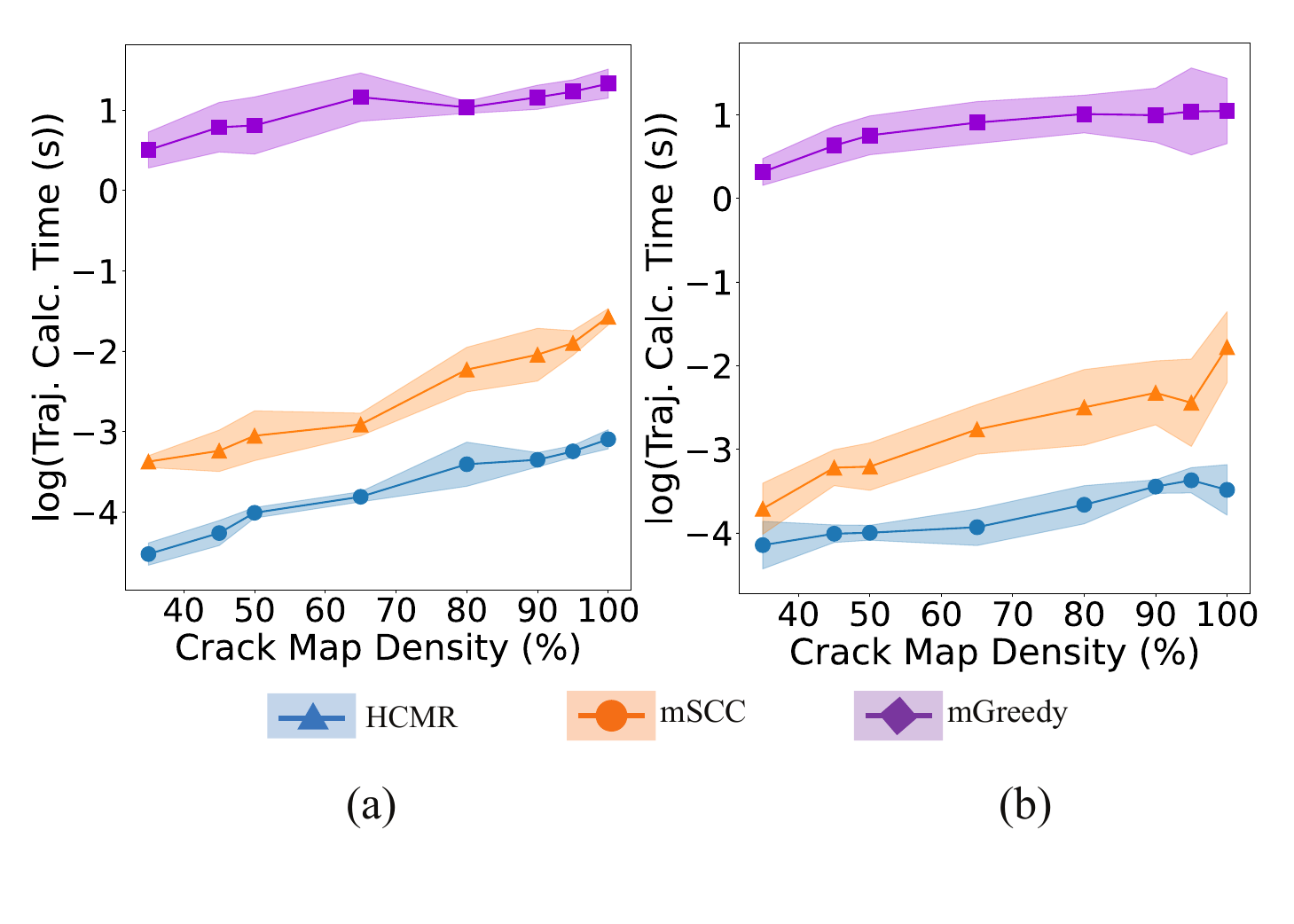}
  
\caption{Comparison of calculation times for trajectory generation using the HCMR, mSCC, and mGreedy planning algorithms. The G-mSCC and S-mSCC methods exhibit identical computational times. The y-axis represents the logarithm (base 10) of the computation time scale. (a) Uniformly distributed cracks. (b) Gaussian-distributed cracks. The lines represent the mean values, while the shaded areas represent one
standard deviation.}
\label{fig:trajectory_statistics}
\end{figure}

To benchmark our results, we implemented three coverage planning algorithms for comparison. \textsc{SCC} was designed to generate a near-optimal solution for the single-robot double coverage problem~\cite{kaiyanTRO2024}, which we further generalize to the multi-robot scenario using two partitioning methods. The first method, denoted as {G-mSCC}, balances partitioning of the Eulerian graph following a similar approach to~\cite{karapetyan2017efficient}. The second method, denoted as {S-mSCC}, splits the workspace into equal parts, similar to~\cite{rekleitis2008efficient}, where each assigned robot performs \textsc{SCC} within its designated region. 
Additionally, we implemented the Greedy algorithm from~\cite{kaiyanTRO2024} as a benchmark, following the same workspace-splitting approach, and denoted it as m{Greedy}.

We used several metrics to evaluate the motion planning,
crack-filling, and multi-robot coordination performance: (1) \textit{total path length}: the total sum of arc lengths traveled by all robots; (2) \textit{task time}: the time taken by the robot with the longest task duration; (3) \textit{sensor coverage}: the percentage of the total sensor-covered areas over the entire work space; (4) \textit{filling coverage}: the percentage of the effectively filled crack length over the total crack length and (5) \textit{sensing conflicts}: the number of spatial intersections between robot trajectories during sensing tasks.

\subsection{Performance Analysis}
We begin by showcasing 16 maps from categories U65, U100, G65, and G100, where three robots perform double coverage tasks using various algorithms, as illustrated in Fig.~\ref{fig:three_robot_demo}. 
According to the simulation outcomes, the planners HCMR and mGreedy successfully covered the free space and completed the crack-filling task on each map, confirming the completeness of these methods. However, mGreedy results in a significantly longer total path and travel time compared to HCMR.   

To further evaluate the planners, we computed several performance metrics and compared the results. Table~\ref{table:three_robot_demo} presents the performance comparison, corresponding to the results in Fig.~\ref{fig:three_robot_demo}. The analysis shows that HCMR consistently outperformed the benchmark planners across all metrics. Notably, HCMR achieved superior sensor coverage and filling coverage, ensuring 100\% crack-filling efficiency. In terms of sensing conflicts, HCMR performed on par with mGreedy, exhibiting zero conflicts.

To further evaluate the algorithms, we conducted simulations to analyze the statistical performance of three robots operating on the dataset introduced above. 
Fig.~\ref{fig:performance_statistics}(a) shows the total path lengths comparison obtained using different algorithms. We also compare with those of a single robot on uniformly distributed and Gaussian-distributed cracks. In general, total path lengths increase linearly with crack map densities, and the total lengths for three robots are consistently lower than those for a single robot. For uniformly distributed cracks, the HCMR-generated paths are consistently shorter than those generated by all other three planners. Additionally, the HCMR-generated paths exhibit minimal growth as crack densities increase. The variance of HCMR paths is the smallest, indicating that HCMR produces more stable trajectories under varying densities. For Gaussian-distributed cracks, it is notable that when crack density is between 35\% and 45\%, the paths generated by HCMR are longer than those generated by G-mSCC. This occurs because, when the crack topology is overly simple, there are insufficient edges in $\mathbb{G}_\text{sim}$ to facilitate both balanced partitioning and total path length optimization. However, as shown in Fig.~\ref{fig:performance_statistics}(b), even in this density range, the average task time using HCMR remains shorter than that using the other three planners.


Regarding task time analysis, similar values are observed at the lowest density. This is because the number of edges in $\mathbb{G}_\text{sim}$ is small, leaving HCMR with limited scope for regulation. In single-robot scenarios, HCMR consistently results in shorter task times, with the difference becoming pronounced as crack density increases. For uniformly distributed cracks, as density increases, the HCMR-generated task time per robot approaches the single-robot task time, with small variance, indicating balanced and stable task assignments among robots. Fig.~\ref{fig:performance_statistics}(c) shows the task time comparison. In contrast, the G-mSCC-generated task times exhibit higher variance and lack the same level of balance, as G-mSCC selects only one random Eulerian tour of $\mathbb{G}_\text{sim}$. Similar tendencies are observed for the other two planners, as well as for scenarios with Gaussian-distributed cracks.

Sensing conflicts is another crucial metric, as robots stopping during high-speed sensing to avoid conflicts is undesirable in practice. Fig.~\ref{fig:performance_statistics}(d) shows that HCMR-generated trajectories exhibit virtually no sensing conflicts, attributed to the continuous Morse invariance. In contrast, other planners, especially G-mSCC-generated trajectories are significantly more prone to conflicts, especially as crack density increases.

Therefore, based on the statistical comparison data depicted in Fig.~\ref{fig:performance_statistics}, the HCMR algorithm significantly improves the planned path length by at least 10.0\%, reduces task time by at least 16.9\% in average, and ensures complete crack coverage and conflict-free operation compared to other benchmark planning methods.

We also conducted simulations with varying numbers of robots, ranging from 1 to 40, on two maps (U100 and G100). The results are presented in Fig.~\ref{fig:robot_num_statistics}. As the number of robots increases, the total path length tends to decrease. For the U100 map, the task time continues to decrease gradually even after $N > 20$, indicating greater potential for equal task assignment. However, for the G100 map, the task time stabilizes after $N > 7$, and the total path length begins to increase after $N > 18$, suggesting the limited capability of equal task assignment on this map. This difference between uniform cracks and Gaussian cracks is also observed in Figs.~\ref{fig:performance_statistics}(b) and (c).

We discuss the computational complexity of the HCMR algorithm. The size of $\mathcal{P}$ depends on the topology of the graph, and its complexity does not have a clear approximation. However, as shown in Fig.~\ref{fig:computation_statistics}, the size of $\mathcal{P}$, the time required to determine $\mathcal{P}$, and the CMS process all grow exponentially with the crack map density. In contrast, the edge sequence BP process does not exhibit significant growth with increasing density and consistently remains within 20~ms.

Even though the calculation of $\mathcal{P}$ exhibits exponential complexity, the local trajectory generation for every path within this collection remains highly efficient, as previously discussed. Fig.~\ref{fig:trajectory_statistics} shows the comparison of computing times among HCMR and other planners. We observe that the trajectory generation time for SCC-based planners and mGreedy is one and five orders of magnitude slower than that of HCMR, respectively. In contrast, HCMR exhibits a much smaller growth rate in computation time, consistently staying within 50~ms. 
This efficiency is crucial because, to determine the optimal trajectory, it is necessary to explore every path in $\mathcal{P}$ and generate the corresponding trajectory.


\section{Conclusion and Limitations}
In this paper, we have discussed the complete and optimal solution for the multi-robot double coverage problem. By projecting the regions induced by the double coverage task onto the manifold space and defining specific operations for topological analysis, we propose an algorithmic framework to achieve complete coverage and obtain optimal routes for multiple robots to efficiently complete the task in a known environment, providing formal proofs for the algorithm. Extensive simulation results confirm the effectiveness of the proposed HCMR algorithm across various target distributions. Furthermore, we compare the HCMR algorithm with other benchmark planning methods. 
The multi-robot simulation results demonstrate that the HCMR algorithm significantly improves planned path length, reduces task completion time, and ensures conflict-free operation compared to state-of-the-art planning methods. 

Our findings highlight that the algorithmic framework effectively optimizes robot trajectories while considering path efficiency, task duration, and collision avoidance. However, simulations suggest that computational complexity increases exponentially with crack density, emphasizing the need for a trade-off algorithm in future implementations. Additionally, the current framework is limited to known environments. As a future direction, we aim to develop online algorithms based on the topological and graph-theoretic foundations established in this work. We are also conducting physical experiments to further validate the proposed planning approach.


\bibliographystyle{plainnat}
\bibliography{ChenRef_AIM}

\end{document}